\documentclass{article}
\usepackage{ijcai23}

\usepackage{times}
\usepackage{soul}
\usepackage{url}
\usepackage[hidelinks]{hyperref}
\usepackage[utf8]{inputenc}
\usepackage[small]{caption}
\usepackage{graphicx}
\usepackage{amsmath}
\usepackage{amsthm}
\usepackage{booktabs}
\usepackage{algorithm}
\usepackage{algorithmic}
\usepackage[switch]{lineno}

\usepackage{multirow}
\usepackage{subcaption}
\usepackage{amsfonts}       
\usepackage{nicefrac}       
\usepackage{microtype}      
\usepackage{xcolor}         
\usepackage[T1]{fontenc}

\hypersetup{colorlinks,citecolor={teal}}

\newcommand\yj[1]{\textcolor{black}{#1}}
\newcommand\jh[1]{\textcolor{black}{#1}}
\newcommand\jhtwo[1]{\textcolor{black}{#1}}
\newcommand\jhcvpr[1]{\textcolor{black}{#1}}
\newcommand\js[1]{\textcolor{black}{#1}}
\newcommand\cam[1]{\textcolor{black}{#1}}
\newcommand\camtwo[1]{\textcolor{black}{#1}}

\DeclareMathOperator*{\argmax}{argmax} 
\DeclareMathOperator*{\argmin}{argmin} 

\title{GeNAS: Neural Architecture Search with Better Generalization}

\iftrue
\author{
Joonhyun Jeong$^{1,2}$\and
Joonsang Yu$^{1,3}$\and
Geondo Park$^2$\and
Dongyoon Han$^3$\And
YoungJoon Yoo$^1$
\affiliations
$^1$NAVER Cloud, ImageVision\\
$^2$KAIST\\
$^3$NAVER AI Lab\\
\emails
\{joonhyun.jeong, joonsang.yu\}@navercorp.com,
geondopark@kaist.ac.kr, \\
\{dongyoon.han, youngjoon.yoo\}@navercorp.com
}
\fi

\begin{document}

\maketitle

\begin{abstract}
    \jhcvpr{
    Neural Architecture Search (NAS) aims to automatically excavate the optimal network architecture with superior test performance. Recent neural architecture search (NAS) approaches rely on validation loss or accuracy to find the superior network for the target data.
   In this paper, we investigate a new neural architecture search measure for excavating architectures with better generalization.}
   We demonstrate that the flatness of the loss surface can be a promising proxy for predicting the generalization capability of neural network architectures.
   We evaluate our proposed method on various search spaces, showing similar or even better performance compared to the state-of-the-art NAS methods. \jhcvpr{Notably, the resultant architecture found by flatness measure generalizes robustly to various shifts in data distribution (e.g. ImageNet-V2,-A,-O), as well as various tasks such as object detection and semantic segmentation.} \cam{Code is available at \url{https://github.com/clovaai/GeNAS}.}
\end{abstract}
\section{Introduction}
 Recently, Neural Architecture Search (NAS)~\cite{liu2018darts,liu2018progressive,hong2022dropnas} has evolved to achieve remarkable accuracy along with the development of human-designed networks \cite{he2016deep,dosovitskiy2020image} on the image recognition task. 
\jhtwo{Several NAS methods \cite{zoph2018learning,chu2020fair,rlnas,liu2018darts,xu2019pc,hong2022dropnas} further have demonstrated generalization ability \yj{(generalizability)} of automatically designed networks with test accuracy and transfer performance onto the other datasets. For the widespread leverage of architectures found by NAS on the various tasks such as object detection~\cite{coco} and segmentation~\cite{cordts2016cityscapes} (task-generalizability), investigating generalizability of each architecture candidate is prerequisite and indispensable.} Despite its importance, quantitative measurement of \yj{generalizability} during the architecture search process is still under-explored. In this paper, we aim to find an optimal proxy measurement to discriminate generalizable architectures during the search process.
 
\begin{figure}[t]
    \centering
    \begin{subfigure}[t]{0.14\textwidth}
        \includegraphics[width=1\columnwidth]{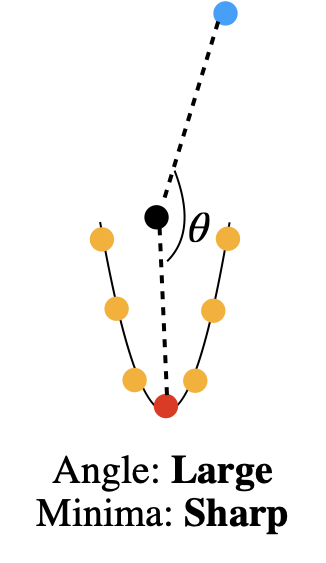}
        \centering
        \caption{ABS}
    \end{subfigure}
    \begin{subfigure}[t]{0.16\textwidth}
        \includegraphics[width=1\columnwidth]{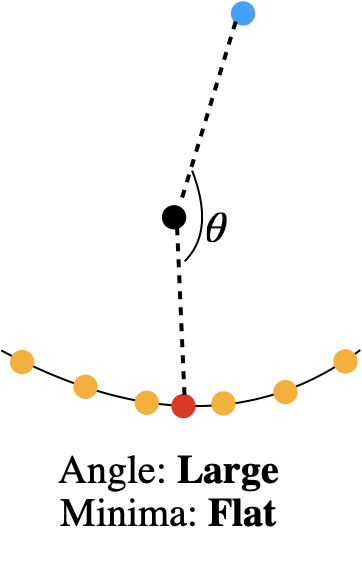}
        \centering
        \caption{FBS (Ours)}
    \end{subfigure}
    \begin{subfigure}[t]{0.14\textwidth}
        \includegraphics[width=1\columnwidth]{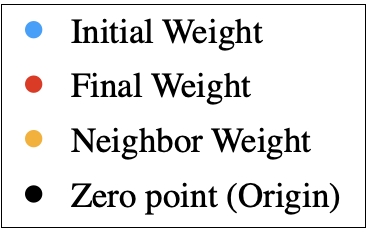}
        \centering
    \end{subfigure}
    \caption{Shape of local loss minima found by angle-based \camtwo{search} (ABS) and flatness-based \camtwo{search} (FBS). (a) \camtwo{The} architecture found by ABS can not guarantee to be located within flat local minima. (b) FBS searches for architectures with flat local minima by inspecting loss values of local neighborhood weights.
    }
    \label{fig:angle_minima_correlation}
\end{figure}

\begin{table}[t]

\centering
\smallskip\noindent
\resizebox{0.85\linewidth}{!}{
\begin{tabular}{c|c|c}
\hline
\multicolumn{3}{c}{Kendall's Tau} \\ \hline
 CIFAR-10 & CIFAR-100 & ImageNet16-120 \\ \hline \hline
 0.4302 & 0.4724 & 0.4097 \\ \hline
\end{tabular}
}
\caption{Low correlation of angle measure with flatness measure on NAS-Bench-201~\protect\cite{nasbench} search space. We evaluated the angle and flatness of all architectures and compared Kendall's Tau \protect\cite{kendall1938new} rank correlation between these search metrics on CIFAR-10, CIFAR-100, and ImageNet16-120 \protect\cite{chrabaszcz} dataset.}
\label{table:nas_bench_201_correlation_btw_metrics}
\end{table}

Previous NAS algorithms including the pioneering differentiable search method, DARTS \cite{liu2018darts} and evolutionary search method, SPOS \cite{spos} use validation performance as a proxy measure for \yj{the generalizability} as follows:
    \begin{equation}
        a^{*} = \argmax_{a \in A} S(a),
    \label{eq:objective}
    \end{equation}
 where $a$ and $A$ denote an architecture candidate and the entire search space, respectively, and \yj{$S(\cdot)$ represents a measurement function which is broadly defined by accuracy} \cite{spos} or negative of loss value \cite{liu2018darts} on a validation dataset.
Although these performance-based \camtwo{search} (PBS) methods find the optimal architecture for generalization on the validation set, they show poor generalizability on the test set and other tasks, caused by overfitting on validation set \cite{zela2019understanding,oymak2021generalization}.
In addition, PBS methods represent a large discrepancy between the validation accuracy and ground truth test accuracy provided by NAS benchmark~\cite{nasbench} as shown in \cite{spos,rlnas}.

To search generalizable architectures, several literatures~\cite{shu2019understanding,rlnas} empirically \yj{observe} that architectures with fast convergence during training have a tendency to show better generalizability on test set. 
Based on the empirical connection between convergence speed and generalization, RLNAS~\cite{rlnas} proposed an Angle-Based \camtwo{Search} (ABS) method, which \yj{estimates} angle \jhcvpr{between initial and final network parameters after convergence of the model (i.e. convergence speed)} as a proxy performance measure during the search process. 
However, we \yj{argue} that ABS still has a large headroom for better generalization in terms of flat (wide) local minima, which has been considered as one of the key signals for inspecting generalizability of a trained network~\cite{keskar2016large,zhang2018deep,pereyra2017regularizing,cha2020cpr,he2019asymmetric}. 
\jhcvpr{
Intuitively, since the architecture with flat loss minima has widely low loss values around the minimum, it can achieve a low generalization error even if the loss surface is shifted due to the distribution gap from the test dataset.
}
Since ABS only concerns the angle between initial model wights and trained ones in terms of convergence speed, found architectures can not be guaranteed to have flat local minima, as shown in Figure \ref{fig:angle_minima_correlation}. Specifically, architectures not \yj{chosen} by ABS (i.e. small angle) might have better generalizability based on the flat property of loss minima. Table \ref{table:nas_bench_201_correlation_btw_metrics} corroborates that angle is indeed in short of correlation with flatness of local minima.

To explicitly design a search proxy measure that has a high correlation with the generalizability of the found model, we propose a flatness-based \camtwo{search} method, namely FBS, which excavates a well-generalizable architecture by measuring the flatness of loss surface. 
FBS can find out robust architecture with low generalization error on shifted data distribution (e.g. test data, \camtwo{out-of-distribution} datasets, downstream tasks) \camtwo{by} inspecting both the depth and flatness of loss curvature near local minima through injecting random noise. In addition, FBS can be either replaced or incorporated with other state-of-the-art search measures to enhance performance as well as generalizability.

Consequently, building upon our search method FBS, we propose a  novel flatness-based NAS framework, namely GeNAS,  to exactly discriminate generalizability of architectures during searching. \yj{We show the effectiveness of the proposed GeNAS for both cases when using FBS solely or integrated into the conventional \jh{architecture} score measurements such as PBS and ABS.}
Specifically, our GeNAS achieves comparable or even better performances on several NAS benchmarks compared to PBS- and ABS-based \camtwo{search} methods \cite{liu2018darts,rlnas,xu2019pc,spos,chu2020fair,chen2019progressive,hong2022dropnas}. 
Furthermore, we also show that the proposed FBS can be combined with conventional search metrics (e.g. PBS, ABS), 
inducing significant performance gain. 
\jhcvpr{Finally, we also demonstrate that our FBS can well-generalize on various data distribution shifts, as well as on multiple downstream tasks such as object detection and semantic segmentation.}
 
 Our contributions can be summarized as follows:
 \begin{itemize}
  
 \item We firstly demonstrate that the flatness of local minima can be employed to quantify generalizability of architecture in NAS domain, which only had been a means of confirming the generalizability after training a neural network.

 \item We propose a new architecture search proxy measure, flatness of local minima, well-suited for finding architectures with better generalization, which can replace or even significantly enhance the \camtwo{search} performance of the existing search proxy measures.

 \item The found architecture induced by our FBS demonstrates the state-of-the-art performance on various search spaces and datasets, \jhcvpr{even showing great robustness on data distribution shift and better generalization on various downstream tasks.}
 
  
\end{itemize}
\section{Related Work}
\subsection{Neural Architecture Search}
 \js{Early NAS methods are based on the reinforcement learning (RL) \cite{baker2016designing,zoph2018learning}, which train the agent network to choose better architecture. The RL-based methods require the test accuracy of each candidate network for reward value, so training every candidate network \jh{from scratch} is also required to measure that. For this reason, it is not feasible on a large-scale dataset such as ImageNet~\cite{krizhevsky2012imagenet}.}
 \js{To solve this problem, the weight-sharing NAS methods are introduced \cite{liu2018darts,xu2019pc,xie2018snas,spos,rlnas}. The weight-sharing NAS generally uses the \textit{SuperNet}, which contains all operations in objective search space, and chooses several operations from the \textit{SuperNet} to decide the final architecture, which is called \textit{SubNet}.}
 Among these weight-sharing NAS frameworks, \cite{liu2018darts,xu2019pc} \camtwo{introduced} a gradient-based \js{architecture search method, where they jointly train the architecture parameters with weight parameters using gradient descent. After training, the final architecture is decided according to the architecture parameters.} 
 Meanwhile, the one-shot NAS methods \cite{spos,bender2018understanding,brock2017smash} pointed out the critical drawback of these gradient-based \camtwo{search} methods as there exists a strongly coupled and biased connection between \textit{SuperNet} weight parameters and its architecture parameters; only a small subset of \textit{SuperNet} weight parameters with large architecture parameter value will be intensely optimized, leaving the others trained insufficiently. 
 Therefore, \cite{spos,bender2018understanding,brock2017smash} sequentially \camtwo{decoupled} the optimization process for \textit{SuperNet} and architecture parameters, showing superior \camtwo{search} performance over the gradient-based \camtwo{search} methods. 
 Inspired by these breakthroughs and its flexibility of introducing various search proxy measures, we construct our GeNAS based on the one-shot NAS framework.

\subsection{Architecture Search Proxy Measure}
\js{During the search time, it is hard to check the actual test performance of each \jh{architecture candidate} when it is trained from scratch, so the proxy measure has to be \jh{employed} for the \jh{candidate} evaluation.}
 Several approaches \camtwo{proposed} to predictively discriminate well-trained neural networks without any training \camtwo{by} inspecting either the correlation between the linear maps of variously augmented image \cite{mellor2021neural} or spectrum of Neural Tangent Kernel (NTK) \cite{chen2021neural}. Although these training-free search proxy measures significantly reduced the search costs within even four GPU hours, actual test performance was inferior to that of training-involved search proxy measures such as validation accuracy and loss.
 Meanwhile, ABS methods \cite{rlnas,hu2020angle} introduced a new search proxy measure, angle, for indicating the generalizability of a neural network architecture, showing search accuracy improvement \cite{rlnas} over conventional search proxy measures such as validation accuracy \cite{spos}. Since ABS method only investigates the convergence speed of an architecture, \cite{rlnas} successfully searched a well-trainable architecture under ground truth label absent during \textit{SuperNet} training. However, searching with randomly-distributed label still shows large performance gap (about 0.15 Kendall's Tau score gap on NAS-Bench-201) to that of searching with the ground-truth label. Therefore, in order to fulfill higher test generalization of a searched architecture, we \camtwo{train} \textit{SuperNet} and searched architecture under ground-truth label setting.
  
\subsection{Flatness of local minima}
 The flatness of loss landscape near local minima has been considered as a key signal for representing generalizability. \cite{keskar2016large,jastrzkebski2017three,hoffer2017train} empirically observed that appropriate training hyper-parameters such as batch size, learning rate, and the number of training iterations can implicitly enable a model to have wide and flat minima, enhancing test generalization performance. \cite{chaudhari2019entropy} further explicitly drives a neural network model to \camtwo{the} flat minima through an entropy-regularized SGD. Several works also promoted \camtwo{the} flat local minima in terms of regularization during training using \textit{Label Smoothing} \cite{pereyra2017regularizing} and \textit{Knowledge Distillation} \cite{zhang2018deep}, enjoying test performance gain. Based on these empirical connections between test generalizability and flatness of local minima of a neural network, we investigate the role of flatness of minima on the architecture search process. \cite{zela2019understanding} has in common with our work in that they also employed a flatness of local minima during the architecture search process, but in an indirect manner. They proposed an early-stopping search process to prevent overfitting on the validation set when the approximated sharpness \jh{of local minima} exceeds the threshold. 
 \jhcvpr{Similarly, \cite{chen2020stabilizing,wang2021neighborhood} tackled to alleviate fluctuating loss surface and accuracy caused by the discretization of architecture parameters in DARTS~\cite{liu2018darts}. We point out these previous similar works lack general usage on various NAS frameworks since they heavily depend on the DARTS~\cite{liu2018darts} framework. Meanwhile, our method can be applied to any architecture search framework without dependence on the architecture parameters of DARTS, such as evolutionary-based \camtwo{search} algorithm~\cite{spos}.
 }
 
    
\section{Method}

\subsection{GeNAS: Generalization-aware NAS with Flatness of local minima}

\jhcvpr{GeNAS is aimed to \jhtwo{search for network architectures with better generalization performance.}} To this end, we introduce a procedure for quantitatively estimating the flatness of an architecture's converged minima as a search proxy measure $F_{val}(\cdot)$ as follows: 
 \begin{equation}
    a^{*} = \argmax_{a \in A} F_{ val}(W_{A}^{*}(a)).
\label{eq:GeNAS_searching}
\end{equation}

 \jhcvpr{Namely, we select the maximal flat architecture $a^{*}$ by evaluating flatness of every \textit{SubNet} extracted from the pre-trained \textit{SuperNet} $W_{A}^{*}$.}
 \yj{From the previous studies~\cite{zhang2018deep,cha2020cpr} that empirically investigated the landscape of converged local minima, the neural networks having flat local minima where the changes of the validation loss around the local minima are relatively small} show better generalization performance at the test phase. Based on these simple but effective empirical connections, we introduce a \jhcvpr{novel method that searches for the architecture with maximal loss flatness around converged minima which can be formulated as below, following \cite{zhang2018deep}:}
\begin{equation}
    F_{ val}(\theta) = (\sum_{i=1}^{t-1}\frac{L(\theta+N(\sigma_{i+1})) - L(\theta+N(\sigma_{i}))}{\sigma_{i+1} - \sigma_{i}})^{-1},
\label{eq:GeNAS_flatness_measure}
\end{equation}
 where $L(\theta)$ denotes validation loss value given weight parameter $\theta$ \yj{abbreviating $W_{A}^{*}(a)$}, and $N(\sigma_i)$ denotes random Gaussian noise with its mean and standard deviation being 0 and $\sigma_i$, respectively. \jhcvpr{Namely, we inspect the flatness of minima near converged weight parameter space by injecting random Gaussian noise. The hyper-parameters $\sigma$ \camtwo{denote} the range for inspection of flatness and $t$ denotes the number of perturbations.}
 \yj{To perturb the weight parameters, we use unidirectional random noise, much \camtwo{more} \jhcvpr{cost-efficient} than recent flatness measuring approaches using Hessian~\cite{yao2019pyhessian} and bidirectional random noise~\cite{he2019asymmetric} which can induce a considerable amount of computational cost.
 We \camtwo{observe} that our choice is sufficient to discriminate  architectures \jhtwo{with high generalization performance}}.

\begin{figure}[t]
    \centering
    \begin{subfigure}[t]{0.9\linewidth}
        \includegraphics[width=1\linewidth]{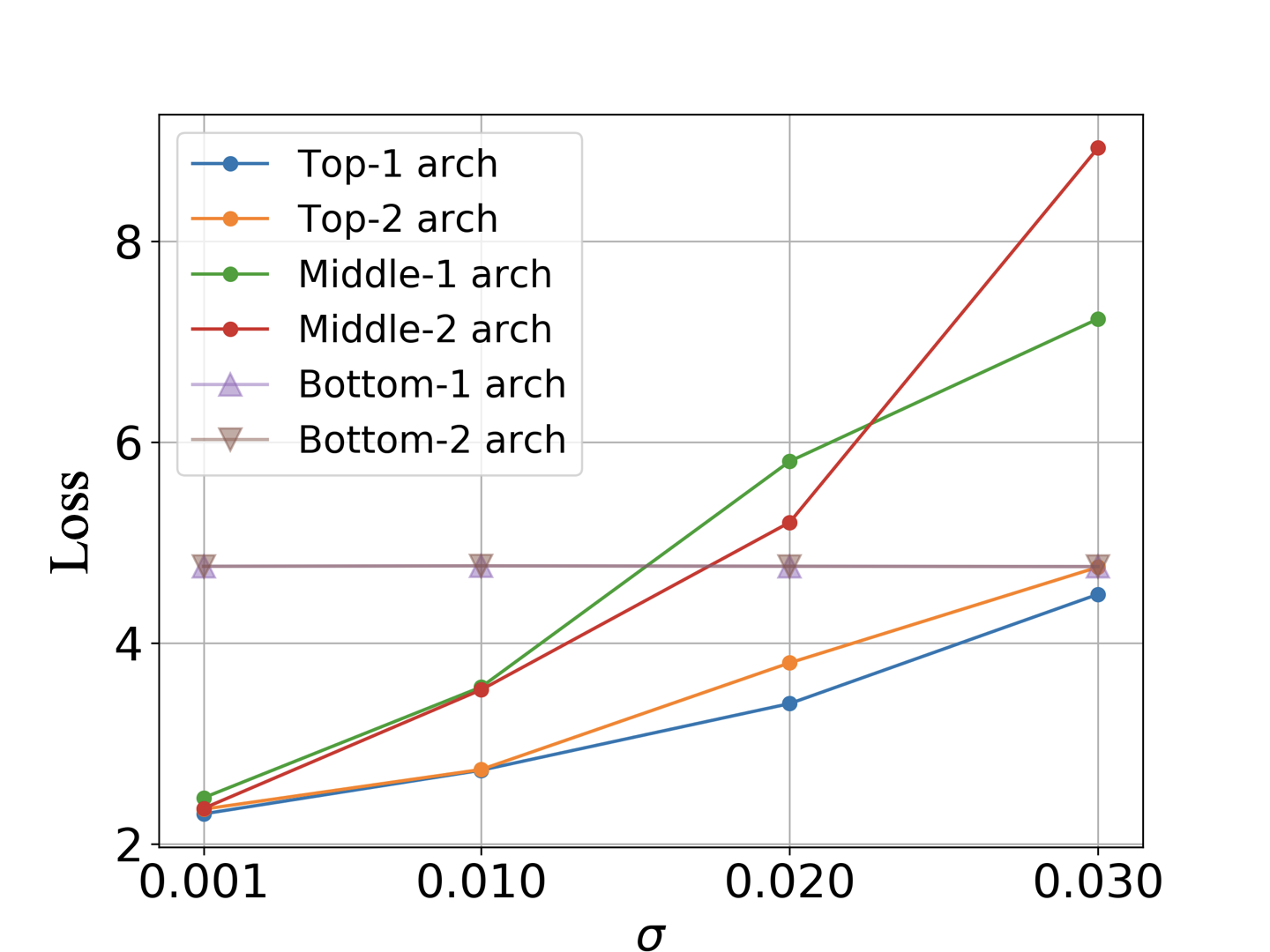}
        \centering
    \end{subfigure}
    \caption{Validation loss curvatures of \jhcvpr{top-k, middle-k, bottom-k} architectures sorted by the ground-truth test accuracy which is given by NAS-Bench-201~\protect\cite{nasbench} on CIFAR-100.}
    \label{fig:loss_landscape_cifar100}
\end{figure}

 Eq \eqref{eq:GeNAS_searching} and \eqref{eq:GeNAS_flatness_measure} would find architecture $a^{*}$ having the flattest local minima in the entire search space, but $a^{*}$ \camtwo{might} have sub-optimal local minima far from the global minimum. In Figure \ref{fig:loss_landscape_cifar100}, bottom-K architectures with the lowest ground test accuracy given by NAS-Bench-201 show the flattest local minima with relatively large loss values compared to the middle-K and top-K architectures. 
 \yj{Therefore, naive investigation of the flatness of an architecture} comes to achieve such sub-optimal architecture in terms of \yj{loss value.} 
 Note that the top-K architectures have the lowest loss values compared to middle and bottom architectures, equipped with flatness near converged minima. \yj{Correspondingly, considering both the flatness of loss landscape and the depth of minima} is essential for excavating a generalizable architecture. \yj{For this reason,} we add an additional term on Eq \eqref{eq:GeNAS_flatness_measure} to search for architectures with deep minima, along with flatness as follows:
 
\begin{equation}
    \begin{aligned}
    F_{val}(\theta) = (\sum_{i=1}^{t-1}|\frac{L(\theta+N(\sigma_{i+1})) - L(\theta+N(\sigma_{i}))}{\sigma_{i+1} - \sigma_{i}}| + \\
    \alpha| \frac{L(\theta+N(\sigma_{1}))}{\sigma_1}|)^{-1}
    \end{aligned}
\label{eq:GeNAS_flatness_measure_with_deep}
\end{equation}
Here, $\sigma_{1}$ denotes the smallest perturbation degree among $\sigma$, hence the second term inspects how low the loss value nearest converged minima is. The term  $\alpha$ denotes the balancing coefficient term between flat and deep minima, which is set to $1$ unless specified.

\subsection{Searching with Combined Metrics}
Recent works~\cite{hosseini2021dsrna,mellor2021neural} adopted a combined search metric for \yj{enhancing the performance of the resultant architecture.} \cite{hosseini2021dsrna} employed an integrated search metric where \yj{the} conventional cross-entropy loss over \yj{a} clean image is combined with approximately measured adversarial robustness lower bound to enhance test accuracy of both clean images and adversarially attacked images. 
Inspired by the weak correlation between \camtwo{the} existing search metrics (e.g. angle) and flatness (Table \ref{table:nas_bench_201_correlation_btw_metrics}), we target to explicitly fulfill the large headroom of conventional search metrics to find better generalizable architectures in terms of our proposed flatness-based search measure (Eq \eqref{eq:GeNAS_flatness_measure_with_deep}). 
Formally, we combine \camtwo{the} existing metrics with flatness as a search proxy measure as follows:
  \begin{equation}
    a^{*} = \argmax_{a \in A} S(W_{A}^{*}(a)) + \gamma\beta F_{val}(W_{A}^{*}(a))
\label{eq:GeNAS_angle_with_flatness}
\end{equation}
where $S$ denotes conventional search metrics such as angle and validation accuracy, $\gamma$ is a balancing \jh{parameter} between \camtwo{the} existing metric and flatness, and $\beta$ is a normalization term, which is fixed as $\sigma_{1}^{-1}$, for matching scale of flatness \jh{term} with \camtwo{the} existing search metric.

\section{Experiments}
\yj{We first} evaluate our proposed GeNAS framework on widely used benchmark dataset, ImageNet with DARTS~\cite{liu2018darts} search space. Furthermore, we thoroughly conduct ablation studies with regard to the components of GeNAS on NAS-Bench-201~\cite{nasbench} benchmark. We refer the reader to the appendix for more experimental details. \jhcvpr{For better confirming robust generalization effect with regard to data distribution shift, we evaluate the found architectures on ImageNet variants (ImageNetV2~\cite{imagenetv2},-A,-O~\cite{imagenet_a_o}). Furthermore, we \camtwo{test} the transferability of our excavated architectures onto other task domains, object detection, and semantic segmentation, with MS-COCO \cite{coco} and Cityscapes~\cite{cordts2016cityscapes} dataset.}

\subsection{ImageNet}

\begin{figure}[t]
    \centering
    \begin{subfigure}[t]{0.9\linewidth}
        \includegraphics[width=1\linewidth]{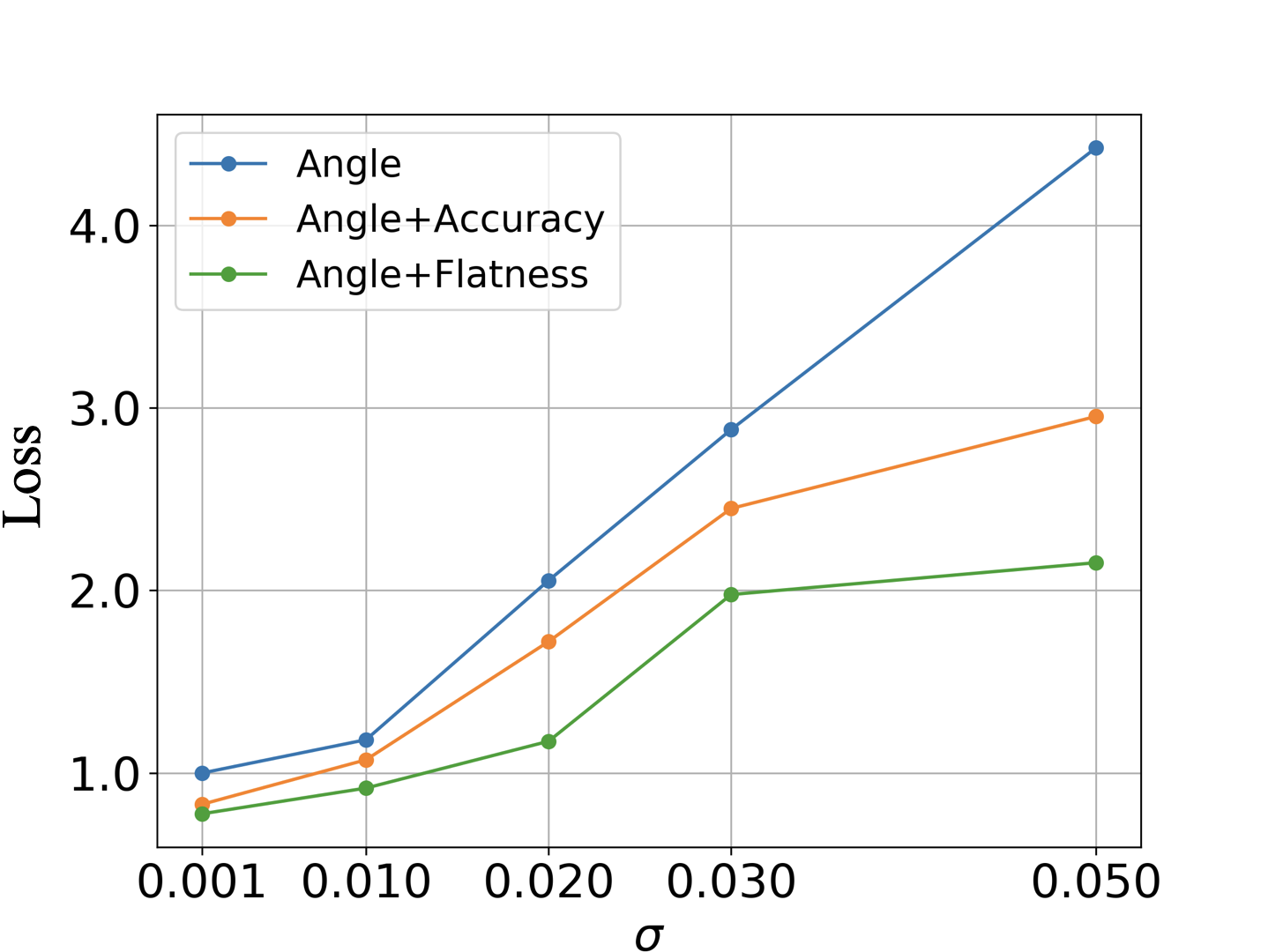}
        \centering
    \end{subfigure}
    \caption{Test loss curvatures of architectures found by \textit{Angle}, \textit{Angle+Accuracy}, \textit{Angle+Flatness}.  
    }
    \label{fig:angle_angleflatness_angleacc_loss_landscape_cifar10}
\end{figure} 


\begin{table*}[t]

\centering
\smallskip\noindent
\resizebox{0.95\linewidth}{!}{
\begin{tabular}{c|c||c|c|c|c}
\hline
Search Dataset & \camtwo{Search} Metric & Params (M) & FLOPs (G) & Top-1 Acc (\%) & Top-5 Acc (\%) \\ \hline  \hline
 \multirow{5}{*}{\begin{tabular}[c]{@{}c@{}}CIFAR-10\end{tabular}} & Angle & 5.3 & 0.59 & 75.70 & 92.45 \\ \cline{2-6}
 & Accuracy & 5.4 & 0.60 & 75.32 & 92.20 \\ \cline{2-6}
 & Flatness & 5.6 & 0.61 & 75.95 & 92.74 \\ \cline{2-6}
 & Angle + Flatness & 5.3 \textcolor{blue}{(+0.0)} & 0.59 \textcolor{blue}{(+0.00)}& 76.06 \textcolor{blue}{(+0.36)}& 92.77 \textcolor{blue}{(+0.32)}\\ \cline{2-6}
 & Accuracy + Flatness & 5.6 \textcolor{blue}{(+0.2)} & 0.61 \textcolor{blue}{(+0.01)}& 75.72 \textcolor{blue}{(+0.40)} & 92.59 \textcolor{blue}{(+0.39)} \\ \hline
 
 \multirow{5}{*}{\begin{tabular}[c]{@{}c@{}}CIFAR-100\end{tabular}} & Angle & 5.4 & 0.61 & 75.00 & 92.31 \\ \cline{2-6}
 & Accuracy & 5.4 & 0.60 & 75.37 & 92.23 \\ \cline{2-6}
 & Flatness & 5.2 & 0.58 & 76.05 & 92.64 \\ \cline{2-6}
 & Angle + Flatness & 5.4 \textcolor{blue}{(+0.0)} & 0.60 \textcolor{blue}{(-0.01)}& 75.72 \textcolor{blue}{(+0.72)}& 92.46 \textcolor{blue}{(+0.15)}\\ \cline{2-6}
 & Accuracy + Flatness & 5.4 \textcolor{blue}{(+0.0)} & 0.60 \textcolor{blue}{(+0.00)}& 75.85 \textcolor{blue}{(+0.48)}& 92.74 \textcolor{blue}{(+0.51)}\\ \hline

 \multirow{5}{*}{\begin{tabular}[c]{@{}c@{}}\cam{ImageNet}\end{tabular}} & \cam{Angle} & 5.4 & 0.60 & 75.09
 & 92.30 \\ \cline{2-6}
 & \cam{Accuracy} & 5.3 & 0.58 & 74.78 & 92.11 \\ \cline{2-6}
 & \cam{Flatness} & 5.3 & 0.59 & 75.49 & 92.38 \\ \cline{2-6}
 & \cam{Angle + Flatness} & 5.5 \textcolor{blue}{(+0.1)} & 0.60 \textcolor{blue}{(+0.00)}& 75.66 \textcolor{blue}{(+0.57)}& 92.62 \textcolor{blue}{(+0.32)}\\ \cline{2-6}
 & \cam{Accuracy + Flatness} & 5.3 \textcolor{blue}{(+0.0)} & 0.59 \textcolor{blue}{(+0.01)}& 75.33 \textcolor{blue}{(+0.55)}& 92.41 \textcolor{blue}{(+0.30)}\\ \hline
 
\end{tabular}
}
\caption{\cam{Performance of various search metrics on ImageNet.}
\jhtwo{The amount of change from adding $Flatness$ term is denoted with blue color.}}
\label{table:cifar10_transfer}
\end{table*}
\subsubsection{Searching on CIFAR-10.}
We analyze the transferability of architectures found on small datasets such as CIFAR-10 and CIFAR-100 onto ImageNet. Specifically, we search architectures with 8 normal cells \cam{(i.e., \textit{stride} = 1)} and 2 reduction cells \cam{(i.e., \textit{stride} = 2)} on CIFAR-10/100, and transfer these normal / reduction cell architectures onto ImageNet by training from scratch and evaluating top-1 accuracy \yj{on ImageNet} validation set.
We compare our proposed FBS with other search metrics on CIFAR-10 in the upper part of Table \ref{table:cifar10_transfer}. As a stand-alone search metric, flatness measure shows the best \camtwo{search} performance among the other metrics \yj{including} accuracy and angle with comparable FLOPs ($\sim=$ 0.6G) and parameters, when transferring searched architecture from CIFAR-10 onto ImageNet. 
Furthermore, when the angle is combined with flatness, \jh{loss landscape of found architecture becomes to be flatter and deeper as shown in Figure \ref{fig:angle_angleflatness_angleacc_loss_landscape_cifar10}. As a result,} \camtwo{search} performance is further improved by 0.36\% top-1 accuracy without any increase of either FLOPs or parameters. \yj{Also, the accuracy-based proxy measure} also \camtwo{achieves} \jh{performance} gain when flatness is combined. The  results show that our proposed flatness search metric indeed serves as a powerful search proxy measure for finding well-transferable architectures \yj{and also enhances} the other search metrics to have a stronger ability to find architectures \jhtwo{with better test generalization performance}.

\begin{table*}[t]

\centering
\smallskip\noindent
\tabcolsep=0.05cm

\resizebox{1\linewidth}{!}{
\begin{tabular}{c|c|c||c|c|c|c}
\hline
Search Dataset & Method & Search Metric & Params (M) & FLOPs (G) & Top-1 Acc (\%) & Top-5 Acc (\%) \\ \hline  \hline
\multirow{9}{*}{\begin{tabular}[c]{@{}c@{}}CIFAR-10\end{tabular}} 
 & DARTS~\cite{liu2018darts} & Val. loss & 4.7 & 0.57 & 73.3 & 91.3 \\ \cline{2-7}
 & PC-DARTS~\cite{xu2019pc} & Val. loss & 5.3 & 0.59 & 74.9 & 92.2 \\ \cline{2-7}
 & FairDARTS-B~\cite{chu2020fair} & Val. loss & 4.8 & 0.54 & 75.1 & 92.5 \\ \cline{2-7}
 & P-DARTS~\cite{chen2019progressive} & Val. loss & 4.9 & 0.56 & 75.6 & 92.6 \\ \cline{2-7}
 & DropNAS$^{\dagger}$~\cite{hong2022dropnas} & Val. loss & 5.4 & 0.60 & 76.0 & 92.8 \\ \cline{2-7}
 & \cam{SANAS}~\cite{hosseini2022saliency} & \cam{Val. loss} & \cam{4.9} & \cam{0.55} & \cam{75.2} & \cam{91.7} \\ \cline{2-7}
 & SPOS~\cite{spos} & Val. acc & 5.4 & 0.60 & 75.3 & 92.2 \\ \cline{2-7}
 & \cam{MF-NAS}~\cite{xue2022max} & \cam{Val. acc} & \cam{4.9} & \cam{0.55} & \cam{75.3} & \cam{-} \\ \cline{2-7}
 & \cam{Shapley-NAS}~\cite{xiao2022shapley} & \cam{Shapley value} & \cam{5.1} & \cam{0.57} & \cam{75.7} & \cam{-} \\ \cline{2-7}
 & RLNAS~\cite{rlnas} & Angle & 5.3 & 0.59 & 75.7 & 92.5 \\ \cline{2-7}
 & SDARTS-RS~\cite{chen2020stabilizing} & Flatness & 5.5 & 0.61 & 75.5 & 92.7 \\ \cline{2-7}
 & SDARTS-ADV~\cite{chen2020stabilizing} & Flatness & 5.5 & 0.62 & 75.6 & 92.4 \\ \cline{2-7}
 & GeNAS (Ours) & Flatness & 5.6 & 0.61 & 76.0 & 92.7 \\ \cline{2-7}
 & GeNAS (Ours) & Angle + Flatness & \textbf{5.3} & \textbf{0.59} & \textbf{76.1} & \textbf{92.8} \\ \hline
 \multirow{7}{*}{\begin{tabular}[c]{@{}c@{}}CIFAR-100\end{tabular}} 
 & PC-DARTS~\cite{xu2019pc} & Val. loss & 5.3 & 0.59 & 74.8 & 92.2 \\ \cline{2-7}
 & DropNAS$^{\dagger}$~\cite{hong2022dropnas} & Val. loss & 5.1 & 0.57 & 75.1 & 92.3 \\ \cline{2-7}
 & P-DARTS~\cite{chen2019progressive} & Val. loss & 5.1 & 0.58 & 75.3 & 92.5 \\ \cline{2-7}
 & SPOS~\cite{spos} & Val. acc & 5.4 & 0.60 & 75.4 & 92.2 \\ \cline{2-7}
 & RLNAS~\cite{rlnas} & Angle & 5.4 & 0.61 & 75.0 & 92.3 \\ \cline{2-7}
 & GeNAS (Ours) & Flatness & \textbf{5.2} & \textbf{0.58} & \textbf{76.1} & \textbf{92.6} \\ \cline{2-7}
 & GeNAS (Ours) & Angle + Flatness & 5.4 & 0.60 & 75.7 & 92.5 \\ \hline
\end{tabular}
}
\caption{ImageNet performance comparison of SOTA NAS methods searched with  DARTS search space on CIFAR-10 and CIFAR-100 dataset. $^{\dagger}$ denotes that SE~\protect\cite{hu2018squeeze} module is excluded for fair comparison with other methods.}
\label{table:cifar10_100_sota}
\end{table*}

\subsubsection{Searching on CIFAR-100.}
\jhcvpr{In middle part of Table \ref{table:cifar10_transfer}, we analyze transferability of architectures found on CIFAR-100 onto ImageNet.}
\yj{The results show that} flatness consistently \yj{reports significantly} superior \camtwo{search} performance even with fewer flops and parameters compared to ABS or PBS metrics, about 1.05\% and 0.68\% better top-1 accuracy, respectively. Furthermore, when flatness is appended onto angle and accuracy as a search proxy measure, top-1 accuracy drastically increases by 0.72\% and 0.48\%, respectively, which was consistently shown in CIFAR-10. 

\vspace{-4mm}
\cam{\subsubsection{Searching on ImageNet.}}

\cam{In the bottom part of Table \ref{table:cifar10_transfer}, we directly search architectures on ImageNet and evaluate validation accuracy on ImageNet to compare in-domain search performance. Similar to the trend of the transfer experiments, our flatness metric \camtwo{achieves} the best \camtwo{search} performance compared to the existing search metrics and improves generalizability of \camtwo{them}.}

\vspace{2mm}
\subsubsection{Comparison with SOTA NAS methods.}

In Table \ref{table:cifar10_100_sota}, our GeNAS clearly represents large headroom compared to the other state-of-the-art NAS methods. Especially in comparison with SDARTS~\cite{chen2020stabilizing} which is a similar approach to GeNAS by using an implicit regularization for smoothing accuracy landscape, our GeNAS outperforms with a comparable number of FLOPs. Table \ref{table:cifar10_transfer} and \ref{table:cifar10_100_sota} results show that our proposed flatness search metric indeed serves as a powerful search proxy measure for finding well-transferable architectures and also enhances the other search metrics to have a stronger ability to find architectures with better test generalization performance.

\subsection{Generalization Ability}
 \jhcvpr{For a more sophisticated investigation of generalization ability, we analyze GeNAS in terms of robustness towards data distribution shift and transferability onto various downstream tasks in Table \ref{table:generalization_ability}.}
 
\subsubsection{Distribution Shift Robustness.}
\jhcvpr{To measure robustness towards data distribution shift, we evaluate our found architectures on ImageNet variants, ImageNet-V2 matched frequency~\cite{imagenetv2} and ImageNet-A~\cite{imagenet_a_o}, where the test-set is distinct from the original ImageNet validation set. The results demonstrate superior robustness compared to the other NAS methods. Our GeNAS widens the performance gap especially when the distribution shift is severe as in ImageNet-A, which has extremely confusing examples.}

\begin{table*}[ht]

\centering
\smallskip\noindent
\tabcolsep=0.1cm
\resizebox{1\linewidth}{!}{
\begin{tabular}{c|c||c|c|c|c|c|c}
\hline

\multirow{2}{*}{\begin{tabular}[c]{@{}c@{}} Method \end{tabular}} & 
\multirow{2}{*}{\begin{tabular}[c]{@{}c@{}} Search \\ Measure \end{tabular}} & 
\multirow{2}{*}{\begin{tabular}[c]{@{}c@{}} Params (M) \end{tabular}} &
\multirow{2}{*}{\begin{tabular}[c]{@{}c@{}} FLOPs (G) \end{tabular}} &
\multirow{2}{*}{\begin{tabular}[c]{@{}c@{}} ImageNet-V2 \\ $Acc$ \end{tabular}} & 
\multirow{2}{*}{\begin{tabular}[c]{@{}c@{}} ImageNet-A \\ $Acc$ \end{tabular}} & 
\multirow{2}{*}{\begin{tabular}[c]{@{}c@{}} COCO \\ $AP$ \end{tabular}} &
\multirow{2}{*}{\begin{tabular}[c]{@{}c@{}} Cityscapes \\ $mIoU$ \end{tabular}} \\ 
& & & & & & \\ \hline \hline

PC-DARTS~\cite{xu2019pc} & Val. loss & 5.3 & 0.59 & 62.53 & 3.85 & 35.56 & 70.68 \\ \hline
DropNAS~\cite{hong2022dropnas} & Val. loss & 5.1 & 0.57 & 63.14 & 4.28 & 36.39 & 71.16 \\ \hline
SPOS~\cite{spos} & Val. acc & 5.4 & 0.60 & 62.84 & 3.91 & 36.04 & 71.70\\ \hline
RLNAS~\cite{rlnas} & Angle & 5.4 & 0.61 & 62.95 & 3.81 & 35.98 & 70.84\\ \hline
\cam{SDARTS-ADV}~\cite{rlnas} & \cam{Flatness} & \cam{5.5} & \cam{0.62} & \cam{62.88} & \cam{4.24} & \cam{36.36} & \cam{71.77}\\ \hline
GeNAS (Ours) & Flatness & 5.2 & 0.58 & \textbf{63.38} & \textbf{5.65} & \textbf{37.05} & \textbf{72.58} \\ \hline
GeNAS (Ours) & Angle + Flatness & 5.4 & 0.60 & 63.32 & 4.37 & 36.59 & 72.05 \\ \hline
\end{tabular}
}
\caption{Comparison with SOTA NAS methods on various ImageNet variants and downstream tasks (object detection with COCO~\protect\cite{coco} and segmentation with Cityscapes~\protect\cite{cordts2016cityscapes}).}
\label{table:generalization_ability}
\end{table*}
\subsubsection{Task Generalization.}
\vspace{3mm}
\noindent\textbf{object detection}
\jhcvpr{
 We evaluate the generalization capability of architectures found by GeNAS on the downstream task, specifically object detection. We firstly re-train architectures found on CIFAR-100 onto ImageNet, and finetune on MS-COCO \cite{coco} dataset. For training, we adopt the default training strategy of RetinaNet \cite{lin2017focal} from Detectron2 \cite{wu2019detectron2}. We only replace the backbone network of RetinaNet for analyzing the sole impact of architectures found by each NAS method. 
 The result shows that our GeNAS framework guided by the flatness measure clearly achieves the best AP scores. In case of RLNAS (angle) combined with flatness as a search metric, AP is enhanced by about \camtwo{$0.61\%$}, without an increase of FLOPs or number of parameters.
 }
 
 \vspace{3mm}
 \noindent\textbf{semantic segmentation}
 \jhcvpr{We also test the generalization of our GeNAS on Semantic Segmentation task with Cityscapes~\cite{cordts2016cityscapes} dataset. Based on the DeepLab-v3~\cite{deeplabv3}, we only replaced the backbone network and trained with MMSegmentation~\cite{mmseg2020} framework. The results demonstrate the effectiveness of our flatness-guided 
 architectures with a large performance margin. Consistently, our flatness guidance ensures a large performance gain, about $1.21\%$, when added onto angle-based \camtwo{search}.}

\subsection{Ablation Study} 

To better analyze our proposed FBS-based GeNAS framework, we conduct an ablation study of each component and hyper-parameters consisting of GeNAS.

\subsubsection{Flatness range.} We analyze the effect of range of inspecting flatness near converged local minima in Table \ref{table:nas_bench_201_ablation_std}. The results demonstrate that searching flat architectures within too small area near converged minima (1st row in Table \ref{table:nas_bench_201_ablation_std}) is not sufficient for discriminating generalizable architectures. When $\sigma$ is increased to $\{2e-3, 1e-2, 2e-2\}$, Kendall's Tau is considerably improved, while further widening the flatness inspection range (4th row in Table \ref{table:nas_bench_201_ablation_std}) only significantly degrades the \camtwo{search} performance on various datasets.

\subsubsection{Deep and low minima.} We further investigate the effect of searching architectures equipped with not only flatness but also the depth of loss landscape near converged minima. Specifically, we adjust $\alpha$ in Eq \eqref{eq:GeNAS_flatness_measure_with_deep}, where $\alpha=0$ denotes searching with only flatness of local minima. Results on Table \ref{table:nas_bench_201_ablation_alpha} demonstrate that as $\alpha$ value increases from \yj{zero to one}, \camtwo{search} performance is drastically enhanced, indicating the indispensability of searching with both flatness and depth of minima. 
Note that $\alpha=0$ case can search out a sub-optimal architecture that has \camtwo{highly} flat loss curvature but its loss values near local minima are too high, as shown in Figure \ref{fig:loss_landscape_cifar100}. When $\alpha$ is further increased to $\alpha > 1$, Kendall's Tau rank correlation starts to decrease, denoting that searching with largely depending on the depth of converged minima is not optimal for discriminating better generalizable architectures.

\subsubsection{Perturbation Methodology.}
   To quantitatively measure flatness of loss landscape, all the weight parameters of a given network are perturbed with random direction following Gaussian distribution as in Eq \eqref{eq:GeNAS_flatness_measure_with_deep}. Here, we investigate the effect of perturbation positions and directions. In Table \ref{table:perturbation}, perturbing only weight parameters of target search cells (i.e. excluding stem \textit{conv} layer and final \textit{fully-connected} layer) only harms Kendall's Tau. Moreover, with regard to the perturbation directions, strongly perturbing the given models' parameters across the hessian eigen-vectors~\cite{yao2019pyhessian} suffers from a slight decrease of Kendall's Tau (Table \ref{table:perturbation}) with large computational overhead induced by approximation of hessian.

\subsubsection{Effect of Flatness on ABS.} 
We analyze the effect of integrating flatness on ABS. Specifically, we adjust $\gamma$ in Eq \eqref{eq:GeNAS_angle_with_flatness}, which balances the coefficient concerning the ratio of flatness to angle term. In Table \ref{table:cifar100_ablation_gamma}, integrating flatness with a small proportion to angle mildly improves top-1 accuracy. As $\gamma$ increases, top-1 accuracy of searched architecture gradually increases to reach 0.72\% improvement over $\gamma=0$ (ABS) case. 

\begin{table}[t]

\centering
\smallskip\noindent
\resizebox{0.95\linewidth}{!}{

\begin{tabular}{c||c|c}
\hline
\multirow{2}{*}{\begin{tabular}[c]{@{}c@{}} $\sigma$ \end{tabular}} & \multicolumn{2}{c}{Kendall's Tau} \\
\cline{2-3}\
 & CIFAR-10 & IN16-120 \\ \hline \hline
 $\{1e-6, 5e-6, 1e-5\}$  & 0.5756 & 0.5524 \\ \hline
 $\{5e-4, 1e-3, 2e-3\}$  & 0.5770 & 0.5531 \\ \hline
 $\{2e-3, 1e-2, 2e-2\}$  & 0.6047 & 0.5800 \\ \hline
 $\{2e-3, 2e-2, 4e-2\}$  & 0.5416 & 0.2364 \\ \hline
 
\end{tabular}
}
\caption{Kendall's Tau on the NAS-Bench-201 search space according to the perturbation range $\sigma$,  \jh{inspecting the effect of flatness range near local minima.} IN16-120 denotes ImageNet16-120 dataset~\protect\cite{nasbench}.}
\label{table:nas_bench_201_ablation_std}
\end{table}
\vspace{2mm}
\begin{table}[t]
\centering
\tabcolsep=0.1cm 
\resizebox{1\linewidth}{!}{%
\begin{tabular}{@{}c|cccccc@{}}
\toprule

& $\alpha = 0$ & $\alpha = 0.1$ & $\alpha = 0.5$ & $\alpha = 1$ & $\alpha=4$ & $\alpha=16$ \\ \midrule
\begin{tabular}[c]{@{}c@{}} Kendall's Tau \\(CIFAR-10)\end{tabular} & 0.1777 & 0.4026 & 0.5890 & 0.6047 & 0.5898 & 0.5820 \\ 
\bottomrule
\end{tabular}
}
\caption{Kendall's Tau on CIFAR-10 with different $\alpha$ in Eq \eqref{eq:GeNAS_flatness_measure_with_deep}.}
\label{table:nas_bench_201_ablation_alpha}
\end{table}

\begin{table}[t]

\centering
\smallskip\noindent
\tabcolsep=0.18cm 
\resizebox{1\linewidth}{!}{
\begin{tabular}{@{}ccc@{}}
\toprule
Perturbation Position & Perturbation Direction & Kendall's Tau \\ \midrule 
 All & Random & 0.6047 \\
 Search Cells & Random & 0.5612 (-0.0435)\\
 All & Hessian & 0.5908 (-0.0139) \\
 \bottomrule
\end{tabular}
}
\caption{Ablation study of perturbation position and direction on CIFAR-10 with NAS-BENCH-201~\protect\cite{nasbench} search space. \textit{All} denotes perturbing all the weight parameters of a given network, while \textit{Search Cells} denotes perturbing only the weight parameters of search cells. The quantities in the parentheses denote the amount of change compared to the default case (first row).}
\label{table:perturbation}
\end{table}
\vspace{-1mm}
\begin{table}[t]

\centering
\smallskip\noindent
\resizebox{1\linewidth}{!}{

\begin{tabular}{c|c||c|c}
\hline
$\gamma$ & $Flatness\;(\%)$ & Top-1 Acc (\%) & Top-5 Acc (\%) \\ \hline  \hline
0 & 0 & 75.00 & 92.31 \\ \hline
0.5 & 20 & 75.22 \textcolor{blue}{(+0.22)} & 92.39 \textcolor{blue}{(+0.08)} \\ \hline
1.5 & 43 & 75.58 \textcolor{blue}{(+0.58)} & 92.44 \textcolor{blue}{(+0.13)} \\ \hline
6 & 76 & 75.63 \textcolor{blue}{(+0.63)} & 92.54 \textcolor{blue}{(+0.23)}\\ \hline
16 & 89 & 75.72 \textcolor{blue}{(+0.72)} & 92.46 \textcolor{blue}{(+0.15)}\\ \hline
\end{tabular}
}
\caption{\camtwo{Search} performance of $Angle + Flatness$ with different $\gamma$ values, where searched on CIFAR-100 and transferred onto ImageNet. $Flatness\;(\%)$ denotes the average ratio of $Flatness$ compared to $Angle$ during evaluation of architectures on evolutionary \camtwo{search} algorithm. \jh{The quantities in the parentheses denote the amount of change compared to the $\gamma = 0$ case.} }
\label{table:cifar100_ablation_gamma}
\end{table}

\subsection{Search Cost Analysis}
\begin{figure}[hbt!]
    \centering
    \includegraphics[width=0.47\textwidth]{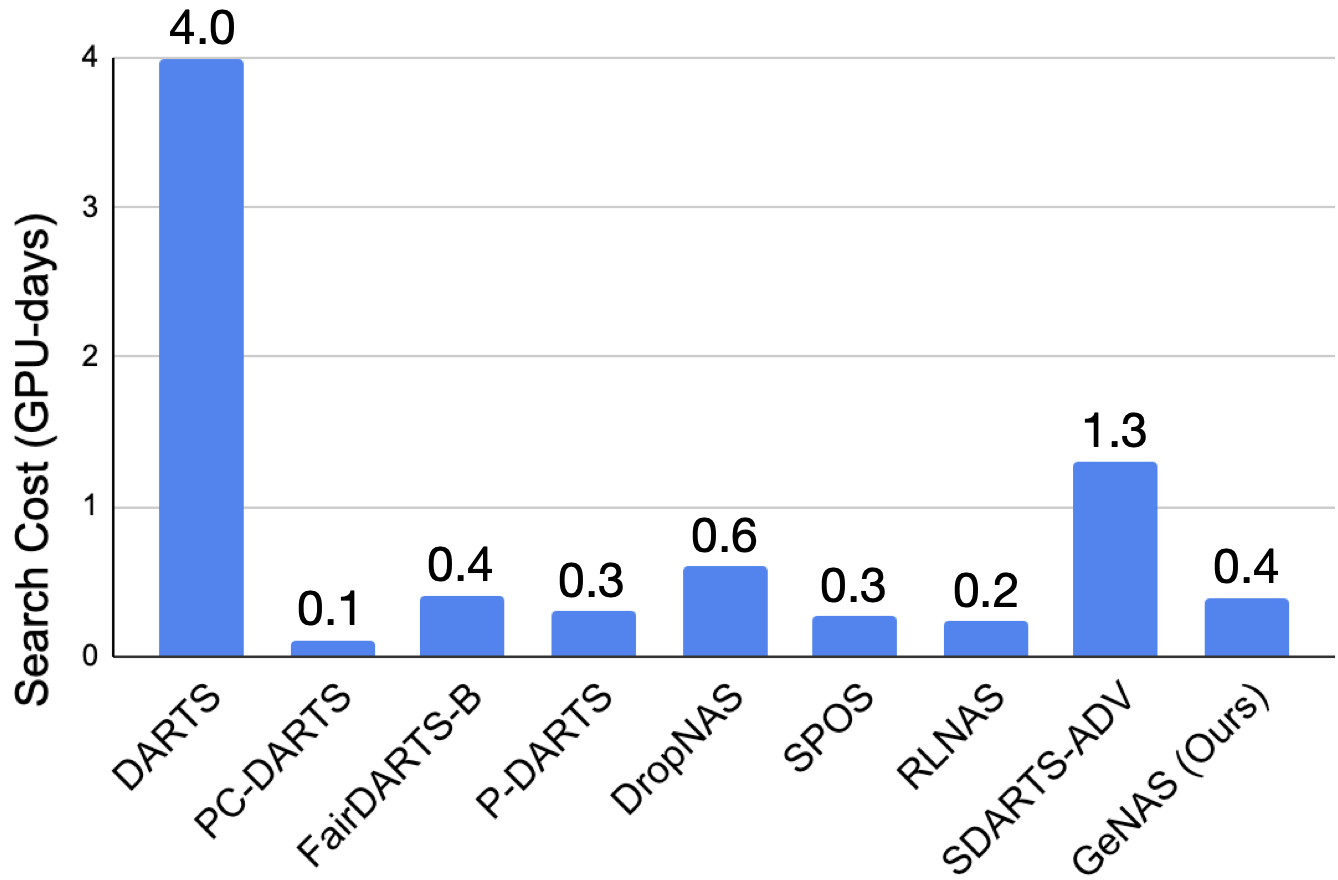}
    \caption{\cam{Comparison of search cost with the SOTA NAS frameworks. 
    }}
    \label{fig:search_cost}
\end{figure}

 \cam{In Figure  \ref{fig:search_cost}, we compare the required architecture search time of GeNAS with the other SOTA NAS frameworks. We measured the execution time spent for the \textit{SuperNet} training and \camtwo{the} search process, using a single NVIDIA V100 GPU. Our required search time is competitive to the other NAS methods while exhibiting shortened time compared to \camtwo{the other} flatness-based \camtwo{search} method (i.e., \camtwo{SDARTS-ADV}).}
 \vspace{3mm}

\section{Conclusion} This paper demonstrates that the flatness of local minima can be directly employed as a proxy of discriminating and searching for generalizable architectures. Based on \camtwo{the} quantitative benchmark experiments on various search spaces and datasets, we demonstrate the superior generalizability of our flatness-based \camtwo{search} over conventional search metrics, while showing comparable or even better \camtwo{search} performance compared to recent state-of-the-art NAS frameworks. We further analyze the insufficient generalizability of conventional search metrics in terms of the flatness of local minima. Consequently, integrating conventional search metrics with our proposed flatness measure can further lead to significantly boosting search performance. \jhcvpr{We also demonstrate superior generalization capability of GeNAS on the downstream object detection and semantic segmentation tasks while showing great robustness with regard to \camtwo{the} data distribution shift.}

\appendix
 \section{Implementation Details}
\label{appendix:train_search_setup}

 \jhcvpr{We construct our GeNAS upon evolutionary-search-based NAS framework~\cite{spos} rather than gradient-based NAS framework~\cite{liu2018darts} because the former one can flexibly embrace a new architecture search proxy measure. 
 The detailed training and searching setups are described as below.}

 \vspace{-5mm}
 \cam{\subsection{SuperNet training and search process.}}
\cam{Following Single Path One-Shot NAS (SPOS)~\cite{spos}, we sequentially optimize the weight parameters of \textit{SuperNet} and select the optimal \textit{SubNet} architecture subsampled from the pre-trained \textit{SuperNet}. Different from SPOS where the optimal \textit{SubNet} is selected by validation accuracy, we leverage the flatness measure to discriminate the well-generalizable \textit{SubNet} architecture. Formally, the entire training and search processes are described as:}

\begin{equation}
W_{A}^{*} =
    \argmin_{W_{A}} \mathbb{E}_{a\sim
U(A)} L_{train}(a, W_A),
\label{eq:SPOS_SuperNet_training}
\end{equation}
\begin{equation}
    a^{*} = \argmax_{a \in A} F_{val}(W_{A}^{*}(a)),
\label{eq:SPOS_searching}
\end{equation}
\cam{where $a$ denotes the \textit{SubNet} architecture inherited from the \textit{SuperNet} architecture $A$, where $W_A$ denotes the weight parameters of \textit{SuperNet}. $W_A(a)$ denotes the weight of the architecture $a$ inherited from the \textit{SuperNet} weight $W_A$. 
$U(A)$ denotes the uniform distribution for sampling $a$ from $A$. In Eq \eqref{eq:SPOS_SuperNet_training}, the \textit{SubNet} weight parameters \yj{$W_A(a)$} selected by random-uniformly are optimized, giving all the \textit{SubNet}s $a\sim
U(A)$ to be \yj{uniformly} optimized. 
In Eq \eqref{eq:SPOS_searching}, given the trained \textit{SuperNet} weight parameters $W_{A}^{*}$, each \textit{SubNet} candidate \yj{$a \in A$} is evaluated by \yj{the architecture score measurement function $F_{val}(\cdot)$}, here defined as flatness of loss surface (i.e., Eq (4) in the manuscript). Consequently, the most flat \textit{SubNet} architecture $a^{*}$ is selected. Following SPOS~\cite{spos}, we conduct the evolutionary search algorithm for Eq \eqref{eq:SPOS_searching}, where top-K populations sorted by flatness repeatedly \textit{Cross-Over} with each other and \textit{Mutate} itself to \yj{search for the architecture having higher flatness value.}}


 \vspace{3mm}

 \subsection{NAS-Bench-201 search space.} NAS-Bench-201 \yj{provides} a relatively small search space where 5 edges with 6 possible operation candidates compose a directed acyclic graph cell, thus \yj{the}  number of architecture candidates from \yj{the} entire search space is $5^6=15625$. 
Using the ground truth test accuracy of all of \yj{the candidate} architectures from NAS-Bench-201, we measure Kendall's Tau score by comparing the rank correlation between search proxy measure and \yj{those from NAS-Bench-201}. 
We use \yj{the equivalent settings to NAS-Bench-201 for constructing} training / validation / test set of CIFAR-10, CIFAR-100, and ImageNet16-120~\cite{chrabaszcz}. 
For training \textit{SuperNet}, we use the same \yj{training} settings (e.g. SGD optimizer with $5e^{-4}$ weight decay factor, 250 training epochs, cosine learning rate scheduling annealed from 0.025 to 0.001) from RLNAS~\cite{rlnas}. During the evolutionary search, we set the entire size of the population as 100 with 20 evolution iterations, following RLNAS. 

\subsection{DARTS search space.}

DARTS~\cite{liu2018darts} has a larger search space than NAS-Bench-201, which provides 8 edges with 7 possible operation candidates (excluding zero operation). 
Furthermore, reduction cell (stride $=$ 2) \cam{which halves the spatial resolution} is also included in the search target, further broadening \yj{the} search space and increasing the difficulty of searching. \cam{We sequentially stack eight normal and two reduction cells where the cells located at 1/3 and 2/3 of the total depth of the network are reduction cells.} We evaluate each \yj{NAS} method by searching architectures on proxy datasets such as CIFAR-10 and CIFAR-100.
\yj{For the selected architectures, we train each model on ImageNet from scratch} and measure the top-1 accuracy. Following RLNAS, we set the number of cells in \textit{SuperNet} as 8 and train 250 epochs. We divide the original training set into training / validation set with equal size on CIFAR-10/100, as in DARTS~\cite{liu2018darts} and PC-DARTS~\cite{xu2019pc}. During the evolutionary search, we set the entire size of the population as 50 with 20 evolution iterations, following SPOS~\cite{spos}. We set $\sigma = \{1e-5, 5e-5, 1e-4\}$, $\{1e-3, 3e-3, 6e-3\}$ for searching on CIFAR-10 and CIFAR-100, respectively. For scratch training on ImageNet, we adjust the initial channels of a target network to have FLOPs around 0.6G. We set the training hyper-parameters exactly the same as PC-DARTS with 8 V100 GPUs.

 \section{Ablation Study}
 
  This section describes an additional ablation study with regard to our proposed Neural Architecture Search (NAS) framework, GeNAS. 
  Specifically, we analyze the effect of flatness-based search (FBS) on accuracy-based search (PBS), along with the effect of PBS on angle-based search (ABS).
  
  \subsection{Effect of FBS on PBS.}
  
  We analyze the effect of integrating our proposed FBS on PBS in Table \ref{table:cifar100_ablation_gamma_accflatness}.
  Integrating flatness with a small proportion shows comparable top-1 and top-5 accuracy compared to PBS ($\gamma = 0$ case). As $\gamma$ increases, top-1 accuracy of searched architecture also increases as to reach $0.48\%$ improvement compared to PBS. 
  
  \subsection{Effect of PBS on ABS.} We further analyze the effect of integrating PBS on ABS in Table \ref{table:cifar100_ablation_gamma_angleacc}. Integrating PBS with a small proportion on ABS improves the top-1 accuracy of ABS. However, as the proportion of PBS increases, top-1 accuracy of searched architecture becomes to be comparable or even degraded compared to that of ABS ($\gamma_{Acc}=0$ case).
  \begin{table}[t]

\centering
\smallskip\noindent
\resizebox{1\linewidth}{!}{
\begin{tabular}{c|c||c|c}
\hline

$\gamma$ & $Flatness\;(\%)$ & Top-1 Acc (\%) & Top-5 Acc (\%) \\ \hline  \hline
0 & 0 & 75.37 & 92.23 \\ \hline
0.25 & 10 & 75.34 (-0.03) & 92.37 \textcolor{blue}{(+0.14)} \\ \hline
2 & 41 & 75.26 (-0.11) & 92.34 \textcolor{blue}{(+0.11)} \\ \hline
8 & 75 & 75.60 \textcolor{blue}{(+0.23)} & 92.36 \textcolor{blue}{(+0.13)} \\ \hline
32 & 92 & 75.85 \textcolor{blue}{(+0.48)} & 92.74 \textcolor{blue}{(+0.51)} \\ \hline
\end{tabular}
}
\caption{Searching performance of $Accuracy + Flatness$ with different $\gamma$ values, where searched on CIFAR-100 and transferred onto ImageNet. $Flatness\;(\%)$ denotes the average ratio of $Flatness$ compared to $Accuracy$ during evaluation of architectures on evolutionary searching algorithm.}
\label{table:cifar100_ablation_gamma_accflatness}
\end{table}
  \begin{table}[t]

\centering
\smallskip\noindent
\resizebox{1\linewidth}{!}{

\begin{tabular}{c|c||c|c}
\hline
$\gamma_{Acc}$ & $Accuracy\;(\%)$ & Top-1 Acc (\%) & Top-5 Acc (\%) \\ \hline  \hline
0 & 0 & 75.00 & 92.31 \\ \hline
0.1 & 12 & 75.32 \textcolor{blue}{(+0.32)} & 92.38 \textcolor{blue}{(+0.07)} \\ \hline
0.5 & 41 & 74.69 (-0.31) & 92.05 (-0.26) \\ \hline
2.5 & 78 & 74.26 (-0.74) & 91.67 (-0.64) \\ \hline
10 & 93 & 75.05 \textcolor{blue}{(+0.05)} & 92.13 (-0.18) \\ \hline
\end{tabular}
}
\caption{Searching performance of $Angle + Accuracy$ with different $\gamma_{Acc}$ values (balancing parameter for $Accuracy$), where searched on CIFAR-100 and transferred onto ImageNet. $Accuracy\;(\%)$ denotes the average ratio of $Accuracy$ compared to $Angle$ during evaluation of architectures on evolutionary searching algorithm. \jh{The quantities in the parentheses denote the amount of change compared to the $\gamma_{Acc} = 0$ case.} }
\label{table:cifar100_ablation_gamma_angleacc}
\end{table}

\begin{figure*}[hbt!]
    \centering
    \begin{subfigure}[t]{0.8\linewidth}
        \includegraphics[width=1\linewidth]{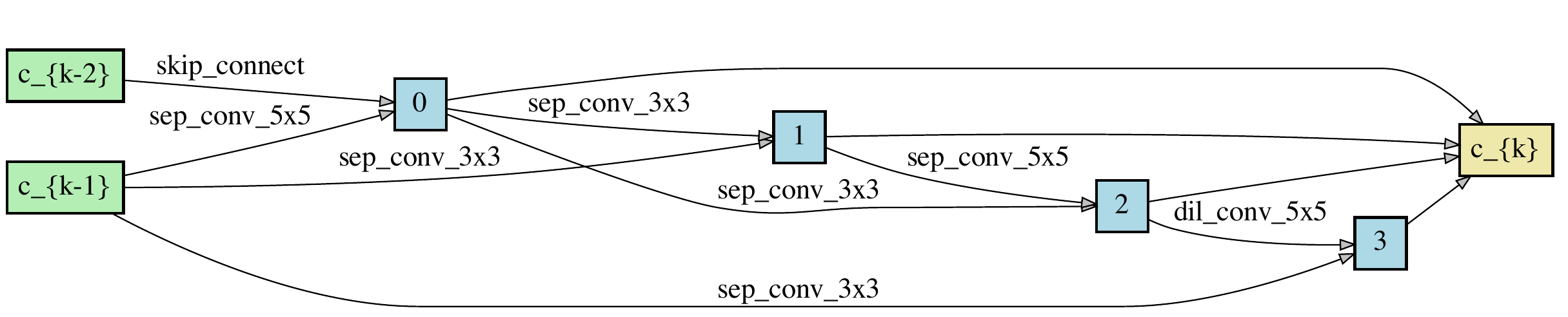}
        \centering
    \subcaption{Normal cell}
    \end{subfigure}
    \begin{subfigure}[t]{0.8\linewidth}
        \includegraphics[width=1\linewidth]{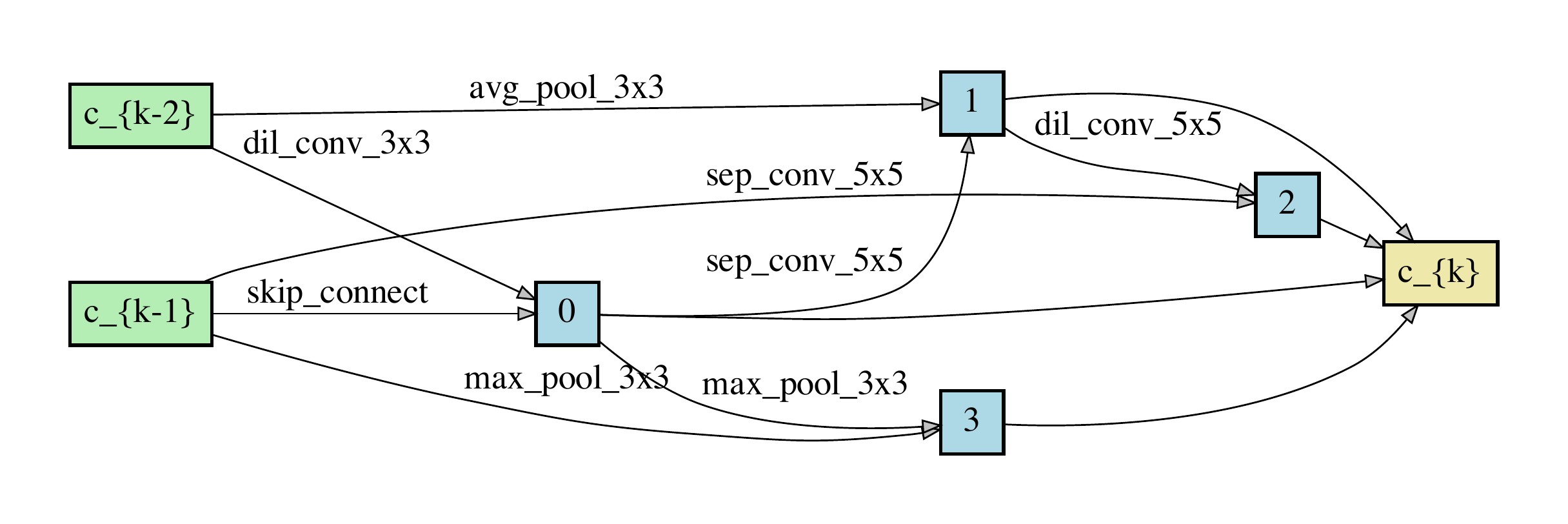}
        \centering
    \subcaption{Reduction cell}
    \end{subfigure}
    \caption{Architecture found by FBS on CIFAR-100 with DARTS~\protect\cite{liu2018darts} search space.}
    \label{fig:fbs_architecture}
\end{figure*}

\begin{figure*}[hbt!]
    \centering
    \begin{subfigure}[t]{0.65\linewidth}
        \includegraphics[width=1\linewidth]{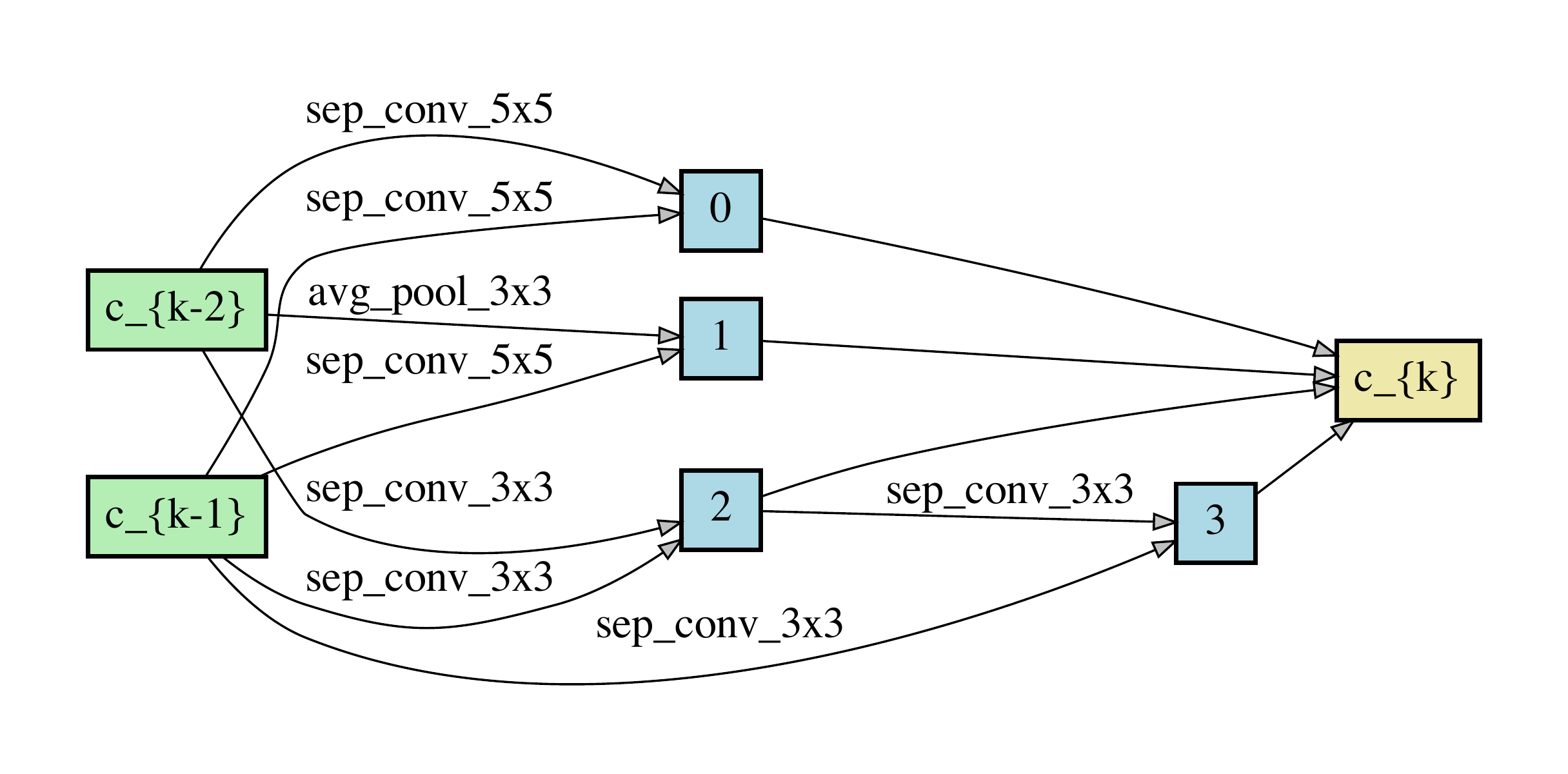}
        \centering
    \subcaption{Normal cell}
    \end{subfigure}
    \begin{subfigure}[t]{0.65\linewidth}
        \includegraphics[width=1\linewidth]{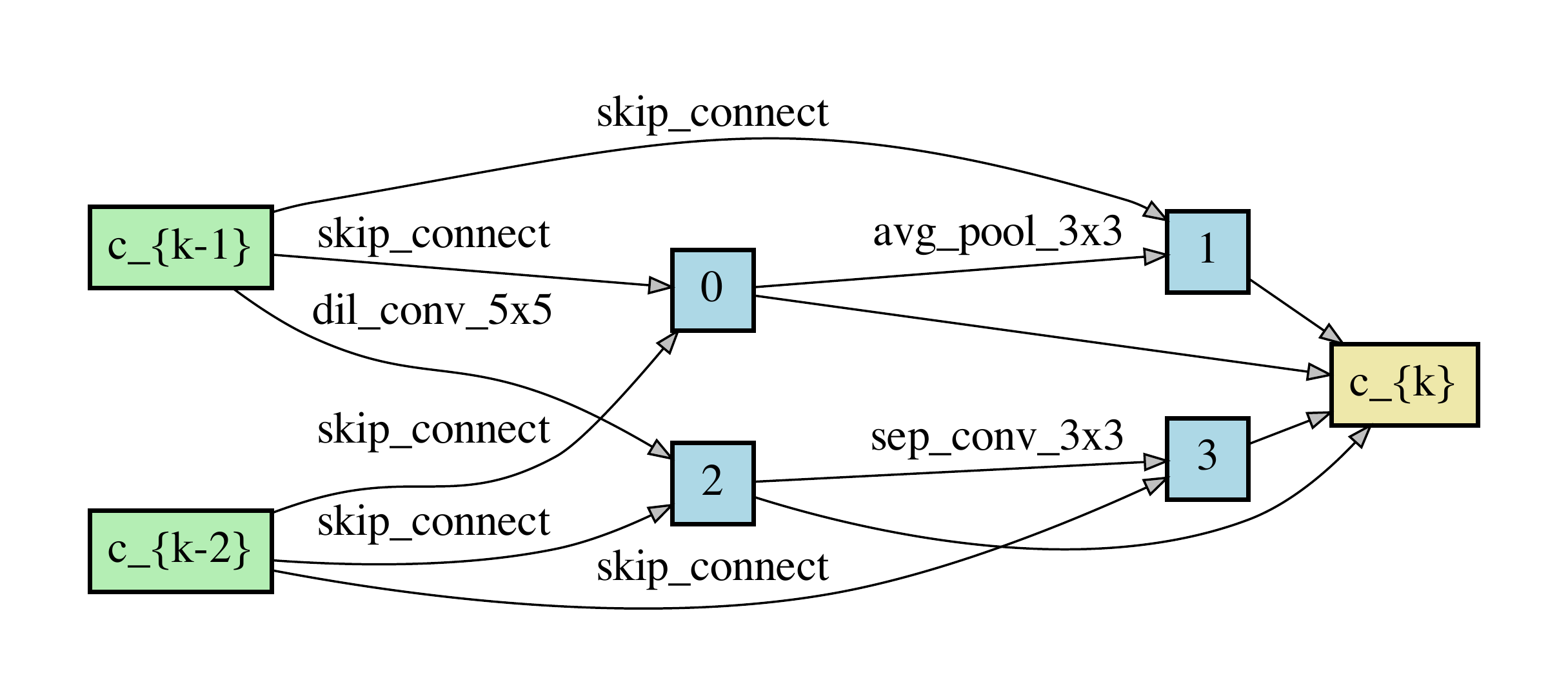}
        \centering
    \subcaption{Reduction cell}
    \end{subfigure}
    \caption{Architecture found by ABS on CIFAR-100 with DARTS search space.}
    \label{fig:abs_architecture}
\end{figure*}

\begin{figure*}[hbt!]
    \centering
    \begin{subfigure}[t]{0.8\linewidth}
        \includegraphics[width=1\linewidth]{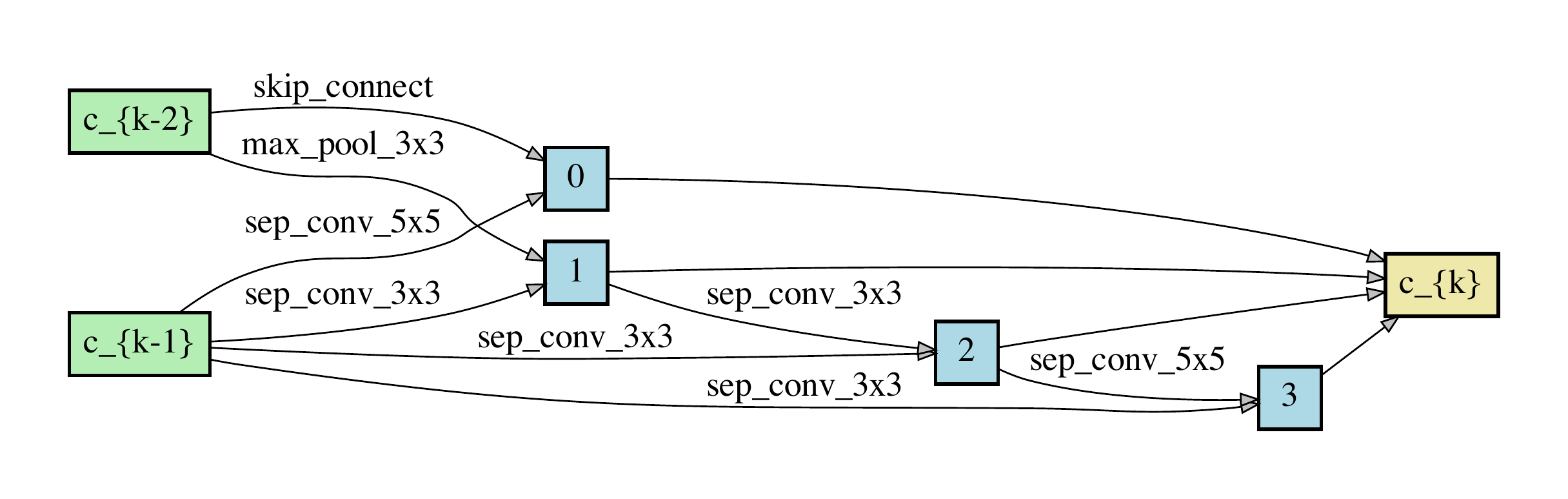}
        \centering
    \subcaption{Normal cell}
    \end{subfigure}
    \begin{subfigure}[t]{0.8\linewidth}
        \includegraphics[width=1\linewidth]{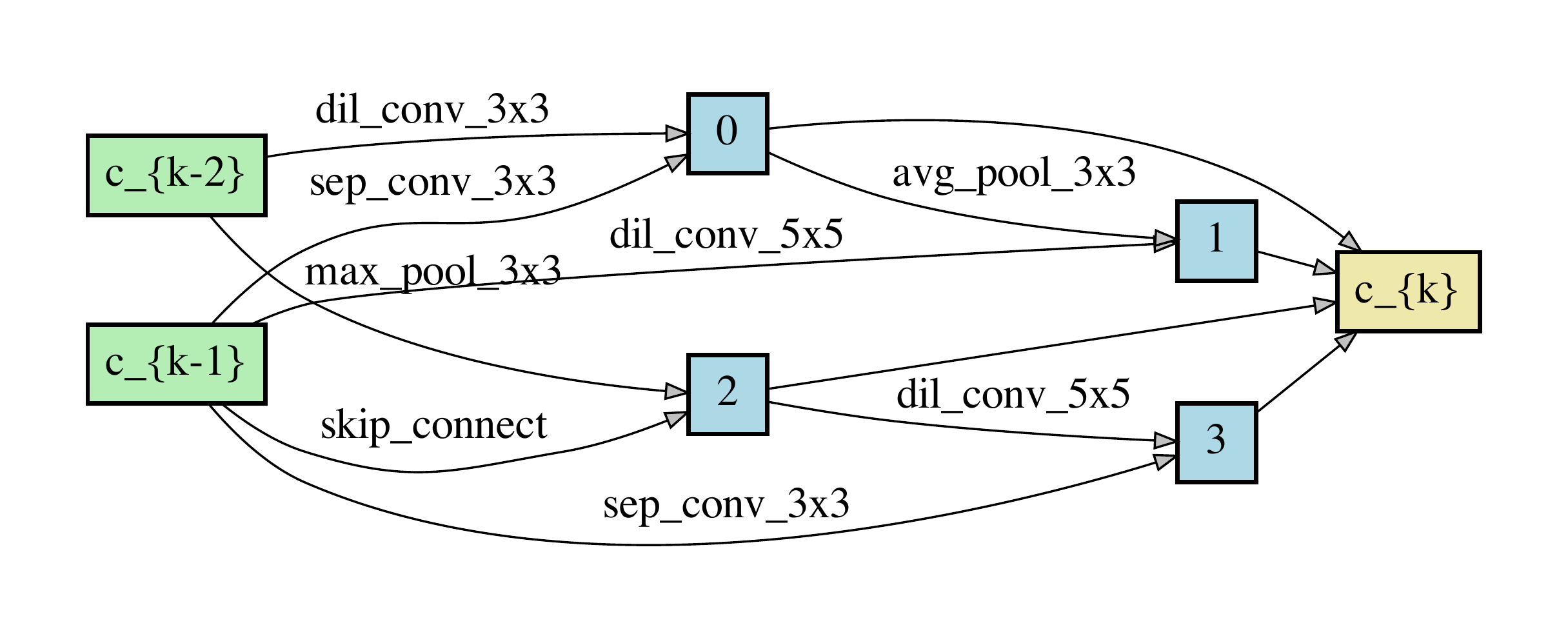}
        \centering
    \subcaption{Reduction cell}
    \end{subfigure}
    \caption{Architecture found by integrating ABS with FBS on CIFAR-100 with DARTS search space.}
    \label{fig:abs_fbs_architecture}
\end{figure*}

  \cam{\section{Visualization of Architectures}}
  \cam{We visualize architecture cells found by our proposed FBS and ABS in Figure \ref{fig:fbs_architecture} and Figure \ref{fig:abs_architecture}, respectively. We further analyze the effect of integrating FBS into ABS with visualization of the resultant architecture cell in Figure \ref{fig:abs_fbs_architecture}. As can be seen in Figure \ref{fig:fbs_architecture} and \ref{fig:abs_architecture}, architecture found by ABS contains several \textit{skip-connect} layers in reduction cell which can possibly lead to sub-optimal architecture as reported in \cite{zela2019understanding}, while that of FBS contains only a single \textit{skip-connect} layer. Moreover, when FBS is integrated into ABS (Figure \ref{fig:abs_fbs_architecture}), the resultant architecture comes to contain fewer skip-connect layers in the reduction cell compared to the ABS case, enjoying less redundancy.}
 \vspace{3mm}
\bibliographystyle{named}
\bibliography{ijcai23}

\begin{thebibliography}{}

\bibitem[\protect\citeauthoryear{Baker \bgroup \em et al.\egroup
  }{2016}]{baker2016designing}
Bowen Baker, Otkrist Gupta, Nikhil Naik, and Ramesh Raskar.
\newblock Designing neural network architectures using reinforcement learning.
\newblock {\em arXiv preprint arXiv:1611.02167}, 2016.

\bibitem[\protect\citeauthoryear{Bender \bgroup \em et al.\egroup
  }{2018}]{bender2018understanding}
Gabriel Bender, Pieter-Jan Kindermans, Barret Zoph, Vijay Vasudevan, and Quoc
  Le.
\newblock Understanding and simplifying one-shot architecture search.
\newblock In {\em International Conference on Machine Learning}, pages
  550--559. PMLR, 2018.

\bibitem[\protect\citeauthoryear{Brock \bgroup \em et al.\egroup
  }{2017}]{brock2017smash}
Andrew Brock, Theodore Lim, James~M Ritchie, and Nick Weston.
\newblock Smash: one-shot model architecture search through hypernetworks.
\newblock {\em arXiv preprint arXiv:1708.05344}, 2017.

\bibitem[\protect\citeauthoryear{Cha \bgroup \em et al.\egroup
  }{2020}]{cha2020cpr}
Sungmin Cha, Hsiang Hsu, Taebaek Hwang, Flavio~P Calmon, and Taesup Moon.
\newblock Cpr: Classifier-projection regularization for continual learning.
\newblock {\em arXiv preprint arXiv:2006.07326}, 2020.

\bibitem[\protect\citeauthoryear{Chaudhari \bgroup \em et al.\egroup
  }{2019}]{chaudhari2019entropy}
Pratik Chaudhari, Anna Choromanska, Stefano Soatto, Yann LeCun, Carlo Baldassi,
  Christian Borgs, Jennifer Chayes, Levent Sagun, and Riccardo Zecchina.
\newblock Entropy-sgd: Biasing gradient descent into wide valleys.
\newblock {\em Journal of Statistical Mechanics: Theory and Experiment},
  2019(12):124018, 2019.

\bibitem[\protect\citeauthoryear{Chen and Hsieh}{2020}]{chen2020stabilizing}
Xiangning Chen and Cho-Jui Hsieh.
\newblock Stabilizing differentiable architecture search via perturbation-based
  regularization.
\newblock In {\em International conference on machine learning}, pages
  1554--1565. PMLR, 2020.

\bibitem[\protect\citeauthoryear{Chen \bgroup \em et al.\egroup
  }{2017}]{deeplabv3}
Liang-Chieh Chen, George Papandreou, Florian Schroff, and Hartwig Adam.
\newblock Rethinking atrous convolution for semantic image segmentation.
\newblock {\em arXiv preprint arXiv:1706.05587}, 2017.

\bibitem[\protect\citeauthoryear{Chen \bgroup \em et al.\egroup
  }{2019}]{chen2019progressive}
Xin Chen, Lingxi Xie, Jun Wu, and Qi~Tian.
\newblock Progressive differentiable architecture search: Bridging the depth
  gap between search and evaluation.
\newblock In {\em Proceedings of the IEEE/CVF International Conference on
  Computer Vision}, pages 1294--1303, 2019.

\bibitem[\protect\citeauthoryear{Chen \bgroup \em et al.\egroup
  }{2021}]{chen2021neural}
Wuyang Chen, Xinyu Gong, and Zhangyang Wang.
\newblock Neural architecture search on imagenet in four gpu hours: A
  theoretically inspired perspective.
\newblock {\em arXiv preprint arXiv:2102.11535}, 2021.

\bibitem[\protect\citeauthoryear{Chrabaszcz \bgroup \em et al.\egroup
  }{2017}]{chrabaszcz}
Patryk Chrabaszcz, Ilya Loshchilov, and Frank Hutter.
\newblock A downsampled variant of imagenet as an alternative to the cifar
  datasets.
\newblock {\em arXiv preprint arXiv:1707.08819}, 2017.

\bibitem[\protect\citeauthoryear{Chu \bgroup \em et al.\egroup
  }{2020}]{chu2020fair}
Xiangxiang Chu, Tianbao Zhou, Bo~Zhang, and Jixiang Li.
\newblock Fair darts: Eliminating unfair advantages in differentiable
  architecture search.
\newblock In {\em European conference on computer vision}, pages 465--480.
  Springer, 2020.

\bibitem[\protect\citeauthoryear{Contributors}{2020}]{mmseg2020}
MMSegmentation Contributors.
\newblock {MMSegmentation}: Openmmlab semantic segmentation toolbox and
  benchmark.
\newblock \url{https://github.com/open-mmlab/mmsegmentation}, 2020.

\bibitem[\protect\citeauthoryear{Cordts \bgroup \em et al.\egroup
  }{2016}]{cordts2016cityscapes}
Marius Cordts, Mohamed Omran, Sebastian Ramos, Timo Rehfeld, Markus Enzweiler,
  Rodrigo Benenson, Uwe Franke, Stefan Roth, and Bernt Schiele.
\newblock The cityscapes dataset for semantic urban scene understanding.
\newblock In {\em Proceedings of the IEEE conference on computer vision and
  pattern recognition}, pages 3213--3223, 2016.

\bibitem[\protect\citeauthoryear{Dong and Yang}{2020}]{nasbench}
Xuanyi Dong and Yi~Yang.
\newblock Nas-bench-201: Extending the scope of reproducible neural
  architecture search.
\newblock {\em arXiv preprint arXiv:2001.00326}, 2020.

\bibitem[\protect\citeauthoryear{Dosovitskiy \bgroup \em et al.\egroup
  }{2020}]{dosovitskiy2020image}
Alexey Dosovitskiy, Lucas Beyer, Alexander Kolesnikov, Dirk Weissenborn,
  Xiaohua Zhai, Thomas Unterthiner, Mostafa Dehghani, Matthias Minderer, Georg
  Heigold, Sylvain Gelly, et~al.
\newblock An image is worth 16x16 words: Transformers for image recognition at
  scale.
\newblock {\em arXiv preprint arXiv:2010.11929}, 2020.

\bibitem[\protect\citeauthoryear{Guo \bgroup \em et al.\egroup }{2020}]{spos}
Zichao Guo, Xiangyu Zhang, Haoyuan Mu, Wen Heng, Zechun Liu, Yichen Wei, and
  Jian Sun.
\newblock Single path one-shot neural architecture search with uniform
  sampling.
\newblock In {\em European Conference on Computer Vision}, pages 544--560.
  Springer, 2020.

\bibitem[\protect\citeauthoryear{He \bgroup \em et al.\egroup
  }{2016}]{he2016deep}
Kaiming He, Xiangyu Zhang, Shaoqing Ren, and Jian Sun.
\newblock Deep residual learning for image recognition.
\newblock In {\em Proceedings of the IEEE conference on computer vision and
  pattern recognition}, pages 770--778, 2016.

\bibitem[\protect\citeauthoryear{He \bgroup \em et al.\egroup
  }{2019}]{he2019asymmetric}
Haowei He, Gao Huang, and Yang Yuan.
\newblock Asymmetric valleys: Beyond sharp and flat local minima.
\newblock {\em Advances in neural information processing systems}, 32, 2019.

\bibitem[\protect\citeauthoryear{Hendrycks \bgroup \em et al.\egroup
  }{2021}]{imagenet_a_o}
Dan Hendrycks, Kevin Zhao, Steven Basart, Jacob Steinhardt, and Dawn Song.
\newblock Natural adversarial examples.
\newblock {\em CVPR}, 2021.

\bibitem[\protect\citeauthoryear{Hoffer \bgroup \em et al.\egroup
  }{2017}]{hoffer2017train}
Elad Hoffer, Itay Hubara, and Daniel Soudry.
\newblock Train longer, generalize better: closing the generalization gap in
  large batch training of neural networks.
\newblock {\em Advances in neural information processing systems}, 30, 2017.

\bibitem[\protect\citeauthoryear{Hong \bgroup \em et al.\egroup
  }{2020}]{hong2022dropnas}
Weijun Hong, Guilin Li, Weinan Zhang, Ruiming Tang, Yunhe Wang, Zhenguo Li, and
  Yong Yu.
\newblock Dropnas: Grouped operation dropout for differentiable architecture
  search.
\newblock In Christian Bessiere, editor, {\em Proceedings of the Twenty-Ninth
  International Joint Conference on Artificial Intelligence, {IJCAI-20}}, pages
  2326--2332. International Joint Conferences on Artificial Intelligence
  Organization, 7 2020.
\newblock Main track.

\bibitem[\protect\citeauthoryear{Hosseini and Xie}{2022}]{hosseini2022saliency}
Ramtin Hosseini and Pengtao Xie.
\newblock Saliency-aware neural architecture search.
\newblock {\em Advances in Neural Information Processing Systems},
  35:14743--14757, 2022.

\bibitem[\protect\citeauthoryear{Hosseini \bgroup \em et al.\egroup
  }{2021}]{hosseini2021dsrna}
Ramtin Hosseini, Xingyi Yang, and Pengtao Xie.
\newblock Dsrna: Differentiable search of robust neural architectures.
\newblock In {\em Proceedings of the IEEE/CVF Conference on Computer Vision and
  Pattern Recognition}, pages 6196--6205, 2021.

\bibitem[\protect\citeauthoryear{Hu \bgroup \em et al.\egroup
  }{2018}]{hu2018squeeze}
Jie Hu, Li~Shen, and Gang Sun.
\newblock Squeeze-and-excitation networks.
\newblock In {\em Proceedings of the IEEE conference on computer vision and
  pattern recognition}, pages 7132--7141, 2018.

\bibitem[\protect\citeauthoryear{Hu \bgroup \em et al.\egroup
  }{2020}]{hu2020angle}
Yiming Hu, Yuding Liang, Zichao Guo, Ruosi Wan, Xiangyu Zhang, Yichen Wei,
  Qingyi Gu, and Jian Sun.
\newblock Angle-based search space shrinking for neural architecture search.
\newblock In {\em European Conference on Computer Vision}, pages 119--134.
  Springer, 2020.

\bibitem[\protect\citeauthoryear{Jastrz{\k{e}}bski \bgroup \em et al.\egroup
  }{2017}]{jastrzkebski2017three}
Stanis{\l}aw Jastrz{\k{e}}bski, Zachary Kenton, Devansh Arpit, Nicolas Ballas,
  Asja Fischer, Yoshua Bengio, and Amos Storkey.
\newblock Three factors influencing minima in sgd.
\newblock {\em arXiv preprint arXiv:1711.04623}, 2017.

\bibitem[\protect\citeauthoryear{Kendall}{1938}]{kendall1938new}
Maurice~G Kendall.
\newblock A new measure of rank correlation.
\newblock {\em Biometrika}, 30(1/2):81--93, 1938.

\bibitem[\protect\citeauthoryear{Keskar \bgroup \em et al.\egroup
  }{2016}]{keskar2016large}
Nitish~Shirish Keskar, Dheevatsa Mudigere, Jorge Nocedal, Mikhail Smelyanskiy,
  and Ping Tak~Peter Tang.
\newblock On large-batch training for deep learning: Generalization gap and
  sharp minima.
\newblock {\em arXiv preprint arXiv:1609.04836}, 2016.

\bibitem[\protect\citeauthoryear{Krizhevsky \bgroup \em et al.\egroup
  }{2012}]{krizhevsky2012imagenet}
Alex Krizhevsky, Ilya Sutskever, and Geoffrey~E Hinton.
\newblock Imagenet classification with deep convolutional neural networks.
\newblock {\em Advances in neural information processing systems}, 25, 2012.

\bibitem[\protect\citeauthoryear{Lin \bgroup \em et al.\egroup }{2014}]{coco}
Tsung-Yi Lin, Michael Maire, Serge Belongie, James Hays, Pietro Perona, Deva
  Ramanan, Piotr Doll{\'a}r, and C~Lawrence Zitnick.
\newblock Microsoft coco: Common objects in context.
\newblock In {\em European conference on computer vision}, pages 740--755.
  Springer, 2014.

\bibitem[\protect\citeauthoryear{Lin \bgroup \em et al.\egroup
  }{2017}]{lin2017focal}
Tsung-Yi Lin, Priya Goyal, Ross Girshick, Kaiming He, and Piotr Doll{\'a}r.
\newblock Focal loss for dense object detection.
\newblock In {\em Proceedings of the IEEE international conference on computer
  vision}, pages 2980--2988, 2017.

\bibitem[\protect\citeauthoryear{Liu \bgroup \em et al.\egroup
  }{2018a}]{liu2018progressive}
Chenxi Liu, Barret Zoph, Maxim Neumann, Jonathon Shlens, Wei Hua, Li-Jia Li,
  Li~Fei-Fei, Alan Yuille, Jonathan Huang, and Kevin Murphy.
\newblock Progressive neural architecture search.
\newblock In {\em Proceedings of the European conference on computer vision
  (ECCV)}, pages 19--34, 2018.

\bibitem[\protect\citeauthoryear{Liu \bgroup \em et al.\egroup
  }{2018b}]{liu2018darts}
Hanxiao Liu, Karen Simonyan, and Yiming Yang.
\newblock Darts: Differentiable architecture search.
\newblock {\em arXiv preprint arXiv:1806.09055}, 2018.

\bibitem[\protect\citeauthoryear{Mellor \bgroup \em et al.\egroup
  }{2021}]{mellor2021neural}
Joe Mellor, Jack Turner, Amos Storkey, and Elliot~J Crowley.
\newblock Neural architecture search without training.
\newblock In {\em International Conference on Machine Learning}, pages
  7588--7598. PMLR, 2021.

\bibitem[\protect\citeauthoryear{Oymak \bgroup \em et al.\egroup
  }{2021}]{oymak2021generalization}
Samet Oymak, Mingchen Li, and Mahdi Soltanolkotabi.
\newblock Generalization guarantees for neural architecture search with
  train-validation split, 2021.

\bibitem[\protect\citeauthoryear{Pereyra \bgroup \em et al.\egroup
  }{2017}]{pereyra2017regularizing}
Gabriel Pereyra, George Tucker, Jan Chorowski, {\L}ukasz Kaiser, and Geoffrey
  Hinton.
\newblock Regularizing neural networks by penalizing confident output
  distributions.
\newblock {\em arXiv preprint arXiv:1701.06548}, 2017.

\bibitem[\protect\citeauthoryear{Recht \bgroup \em et al.\egroup
  }{2019}]{imagenetv2}
Benjamin Recht, Rebecca Roelofs, Ludwig Schmidt, and Vaishaal Shankar.
\newblock Do imagenet classifiers generalize to imagenet?
\newblock In {\em International Conference on Machine Learning}, pages
  5389--5400. PMLR, 2019.

\bibitem[\protect\citeauthoryear{Shu \bgroup \em et al.\egroup
  }{2019}]{shu2019understanding}
Yao Shu, Wei Wang, and Shaofeng Cai.
\newblock Understanding architectures learnt by cell-based neural architecture
  search.
\newblock {\em arXiv preprint arXiv:1909.09569}, 2019.

\bibitem[\protect\citeauthoryear{Wang \bgroup \em et al.\egroup
  }{2021}]{wang2021neighborhood}
Xiaofang Wang, Shengcao Cao, Mengtian Li, and Kris~M Kitani.
\newblock Neighborhood-aware neural architecture search.
\newblock {\em arXiv preprint arXiv:2105.06369}, 2021.

\bibitem[\protect\citeauthoryear{Wu \bgroup \em et al.\egroup
  }{2019}]{wu2019detectron2}
Yuxin Wu, Alexander Kirillov, Francisco Massa, Wan-Yen Lo, and Ross Girshick.
\newblock Detectron2.
\newblock \url{https://github.com/facebookresearch/detectron2}, 2019.

\bibitem[\protect\citeauthoryear{Xiao \bgroup \em et al.\egroup
  }{2022}]{xiao2022shapley}
Han Xiao, Ziwei Wang, Zheng Zhu, Jie Zhou, and Jiwen Lu.
\newblock Shapley-nas: Discovering operation contribution for neural
  architecture search.
\newblock In {\em Proceedings of the IEEE/CVF Conference on Computer Vision and
  Pattern Recognition}, pages 11892--11901, 2022.

\bibitem[\protect\citeauthoryear{Xie \bgroup \em et al.\egroup
  }{2018}]{xie2018snas}
Sirui Xie, Hehui Zheng, Chunxiao Liu, and Liang Lin.
\newblock Snas: stochastic neural architecture search.
\newblock {\em arXiv preprint arXiv:1812.09926}, 2018.

\bibitem[\protect\citeauthoryear{Xu \bgroup \em et al.\egroup
  }{2019}]{xu2019pc}
Yuhui Xu, Lingxi Xie, Xiaopeng Zhang, Xin Chen, Guo-Jun Qi, Qi~Tian, and
  Hongkai Xiong.
\newblock Pc-darts: Partial channel connections for memory-efficient
  architecture search.
\newblock {\em arXiv preprint arXiv:1907.05737}, 2019.

\bibitem[\protect\citeauthoryear{Xue \bgroup \em et al.\egroup
  }{2022}]{xue2022max}
Chao Xue, Xiaoxing Wang, Junchi Yan, and Chun-Guang Li.
\newblock A max-flow based approach for neural architecture search.
\newblock In {\em Computer Vision--ECCV 2022: 17th European Conference, Tel
  Aviv, Israel, October 23--27, 2022, Proceedings, Part XX}, pages 685--701.
  Springer, 2022.

\bibitem[\protect\citeauthoryear{Yao \bgroup \em et al.\egroup
  }{2019}]{yao2019pyhessian}
Zhewei Yao, Amir Gholami, Kurt Keutzer, and Michael Mahoney.
\newblock Pyhessian: Neural networks through the lens of the hessian.
\newblock {\em arXiv preprint arXiv:1912.07145}, 2019.

\bibitem[\protect\citeauthoryear{Zela \bgroup \em et al.\egroup
  }{2019}]{zela2019understanding}
Arber Zela, Thomas Elsken, Tonmoy Saikia, Yassine Marrakchi, Thomas Brox, and
  Frank Hutter.
\newblock Understanding and robustifying differentiable architecture search.
\newblock {\em arXiv preprint arXiv:1909.09656}, 2019.

\bibitem[\protect\citeauthoryear{Zhang \bgroup \em et al.\egroup
  }{2018}]{zhang2018deep}
Ying Zhang, Tao Xiang, Timothy~M Hospedales, and Huchuan Lu.
\newblock Deep mutual learning.
\newblock In {\em Proceedings of the IEEE conference on computer vision and
  pattern recognition}, pages 4320--4328, 2018.

\bibitem[\protect\citeauthoryear{Zhang \bgroup \em et al.\egroup
  }{2021}]{rlnas}
Xuanyang Zhang, Pengfei Hou, Xiangyu Zhang, and Jian Sun.
\newblock Neural architecture search with random labels.
\newblock In {\em Proceedings of the IEEE/CVF Conference on Computer Vision and
  Pattern Recognition}, pages 10907--10916, 2021.

\bibitem[\protect\citeauthoryear{Zoph \bgroup \em et al.\egroup
  }{2018}]{zoph2018learning}
Barret Zoph, Vijay Vasudevan, Jonathon Shlens, and Quoc~V Le.
\newblock Learning transferable architectures for scalable image recognition.
\newblock In {\em Proceedings of the IEEE conference on computer vision and
  pattern recognition}, pages 8697--8710, 2018.

\end{thebibliography}

\end{document}